\documentclass[10pt,twocolumn,letterpaper]{article}

\usepackage{wacv}
\usepackage{times}
\usepackage{epsfig}
\usepackage{graphicx}
\usepackage{amsmath}
\usepackage{amssymb}
\usepackage{booktabs}

\usepackage{algorithm}
\usepackage{algorithmic}
\usepackage{url}
\usepackage{subcaption}
\usepackage{graphicx}
\usepackage{bbm}
\usepackage{tikz}
\usepackage{pgfplots}
\usepackage{xcolor}
\usepackage[normalem]{ulem}

\newcommand{\bfI}{\mathbf{I}}

\newcommand{\bfF}{\mathbf{F}}

\newcommand{\bfP}{\mathbf{P}}

\newcommand{\bfM}{\mathbf{M}}

\usepackage{multirow}

\wacvalgorithmstrack   

\wacvfinalcopy

\ifwacvfinal
\usepackage[breaklinks=true,bookmarks=false]{hyperref}
\else
\usepackage[pagebackref=true,breaklinks=true,colorlinks,bookmarks=false]{hyperref}
\fi

\begin{document}

\title{Unsupervised Image Semantic Segmentation through Superpixels and Graph Neural Networks}

\author{Moshe Eliasof\\
Ben-Gurion University of the Negev\\
Beer-Sheva, Israel\\
{\tt\small eliasof@post.bgu.ac.il}
\and
Nir Ben Zikri\\
Ben-Gurion University of the Negev\\
Beer-Sheva, Israel\\
{\tt\small nirbenz@post.bgu.ac.il}
\and
Eran Treister\\
Ben-Gurion University of the Negev\\
Beer-Sheva, Israel\\
{\tt\small erant@cs.bgu.ac.il}
}

\maketitle
\thispagestyle{empty}

\begin{abstract}
Unsupervised image segmentation is an important task in many real-world scenarios where labelled data is of scarce availability. In this paper we propose a novel approach that harnesses recent advances in unsupervised learning using a combination of Mutual Information Maximization (MIM), Neural Superpixel Segmentation and Graph Neural Networks (GNNs) in an end-to-end manner, an approach that has not been explored yet. We take advantage of the compact representation of superpixels and combine it with GNNs in order to learn strong and semantically meaningful representations of images. Specifically, we show that our GNN based approach allows to model interactions between distant pixels in the image and serves as a strong prior to existing CNNs for an improved accuracy. Our experiments reveal both the qualitative and quantitative advantages of our approach compared to current state-of-the-art methods over four popular datasets.
\end{abstract}


\section{Introduction}
\label{sec:intro}

\begin{figure}
    \centering
    \begin{subfigure}{0.115\textwidth}
    \includegraphics[width=1\textwidth]
    {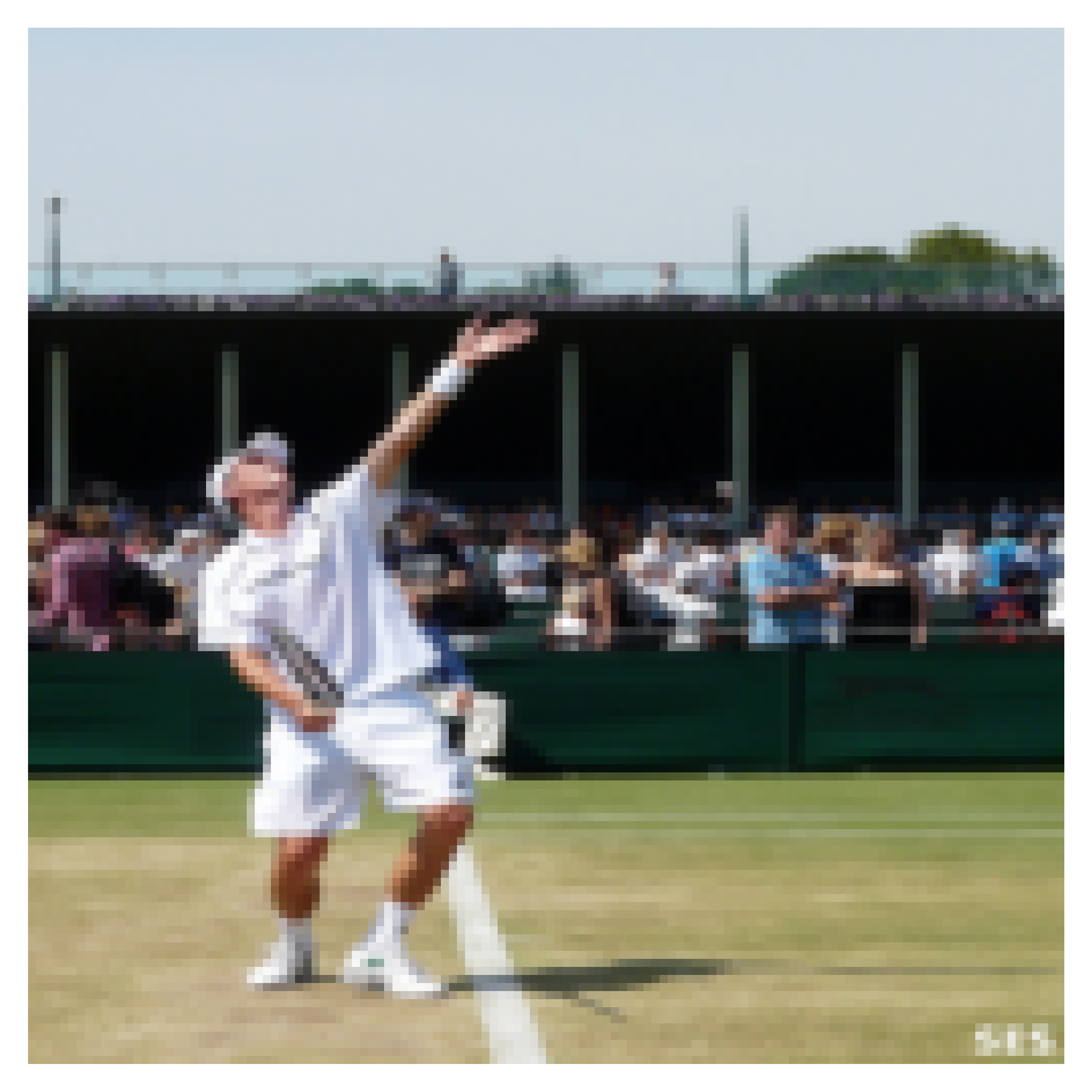} \\
        \includegraphics[width=1\textwidth]
    {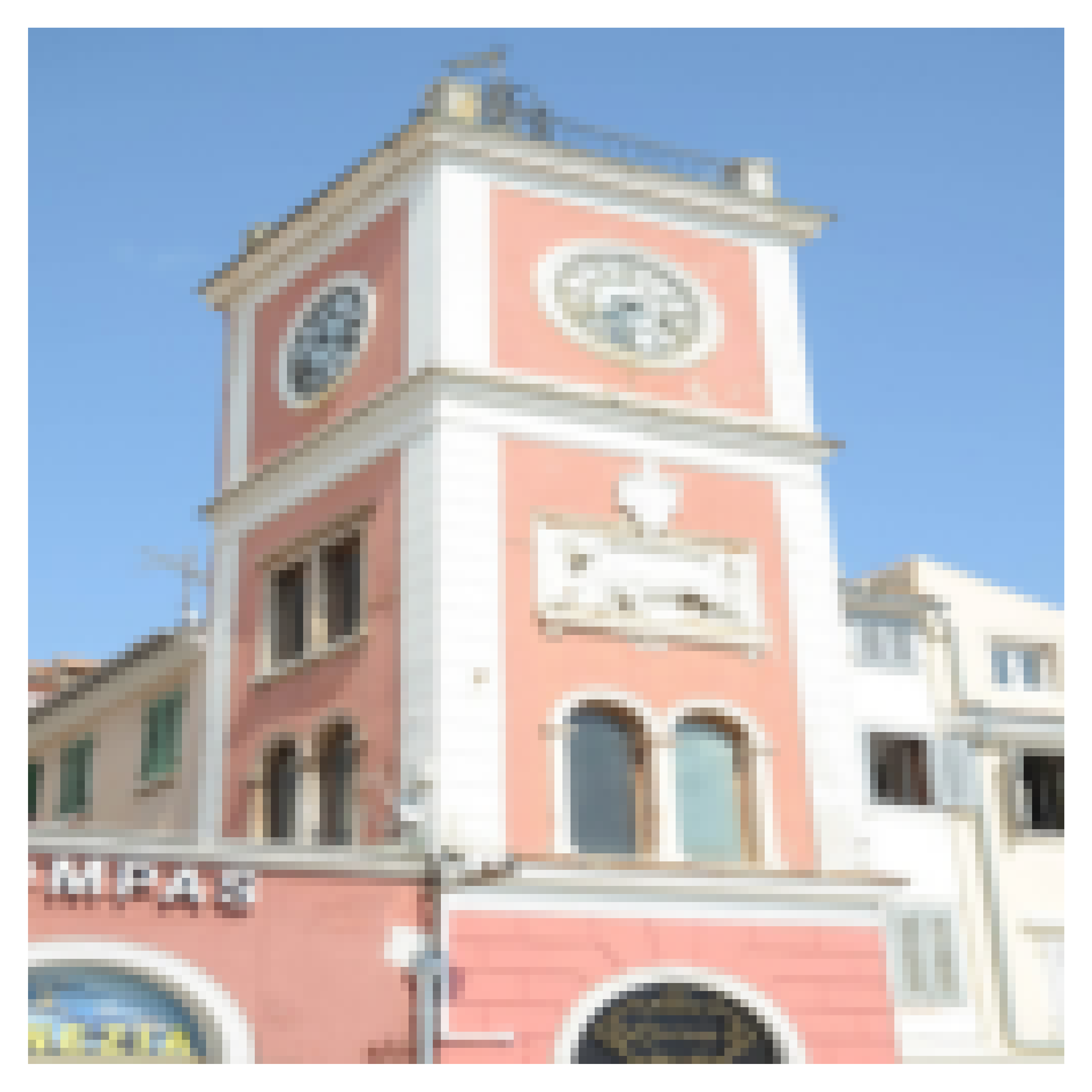} \\
        \includegraphics[width=1\textwidth]
    {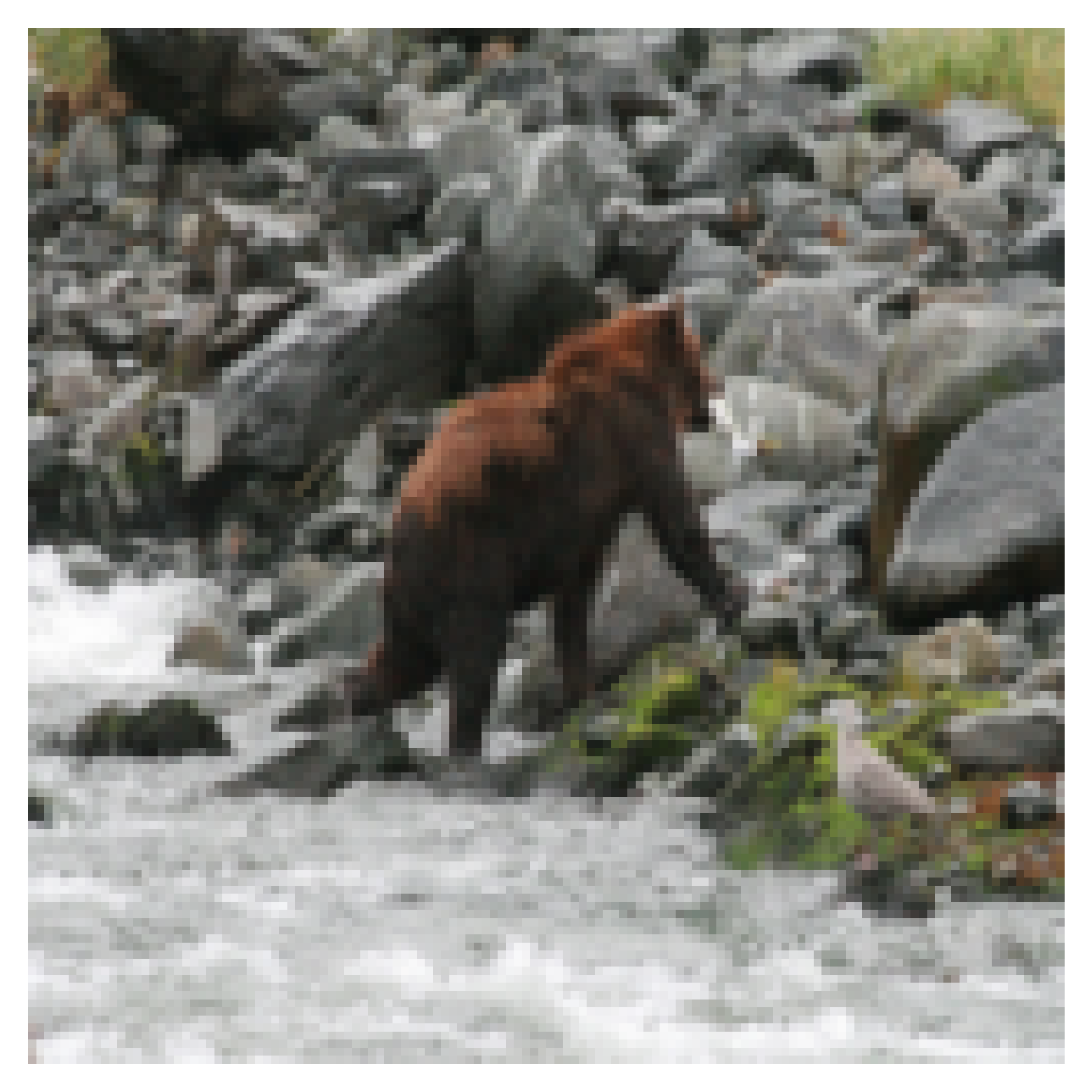}
    \caption{}
    \end{subfigure}
    \begin{subfigure}{0.115\textwidth}
    \includegraphics[width=1\textwidth]{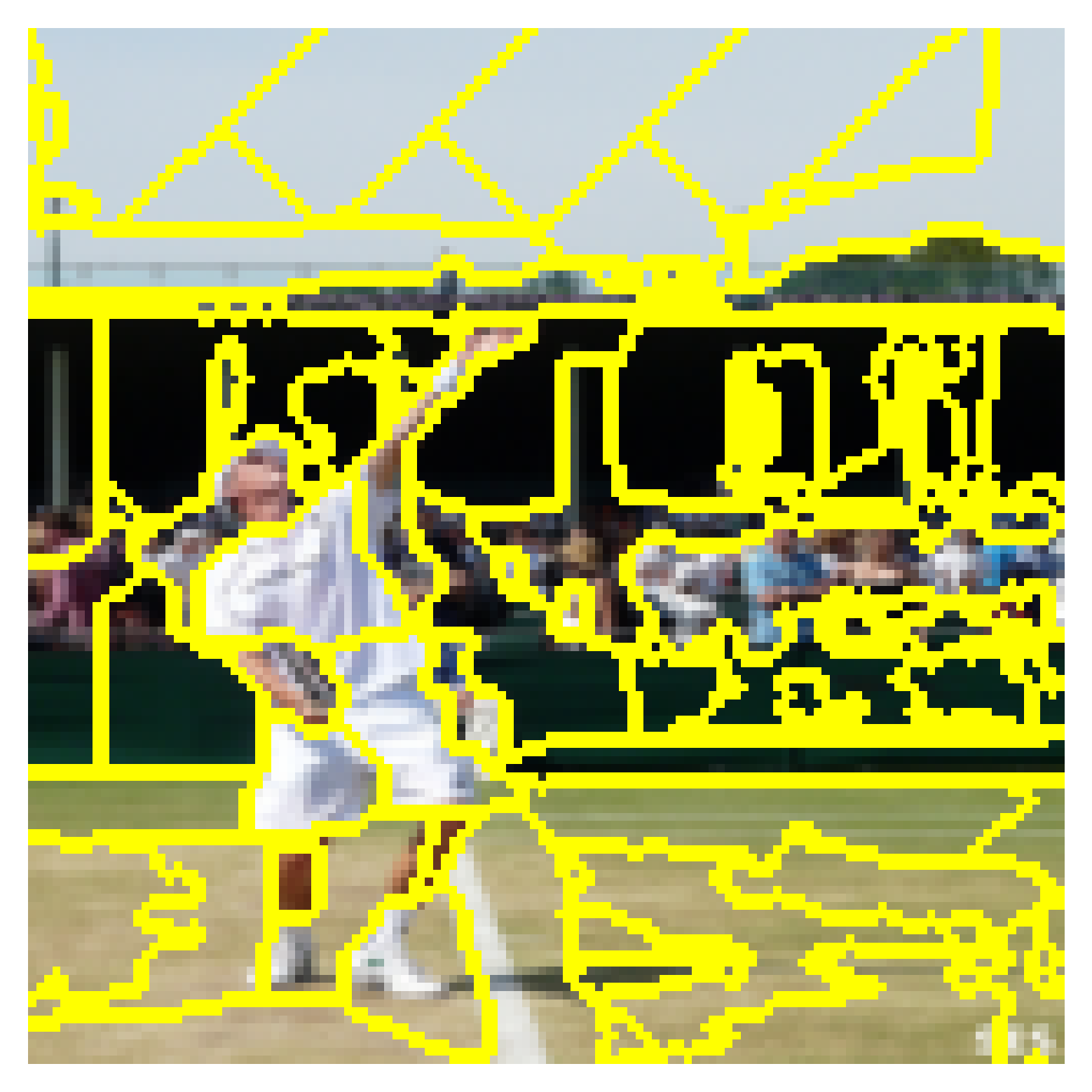} \\
    \includegraphics[width=1\textwidth]{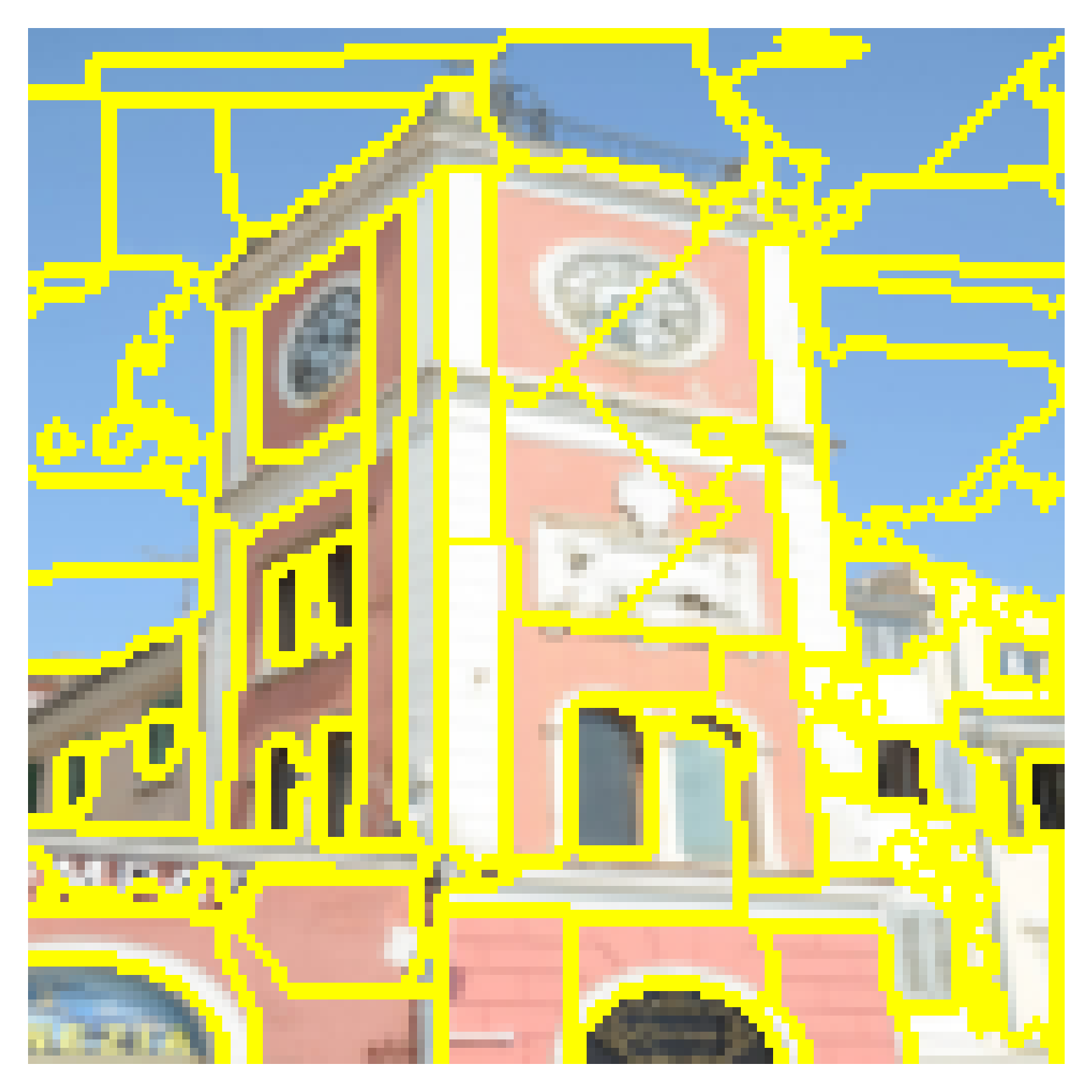} \\
    \includegraphics[width=1\textwidth]{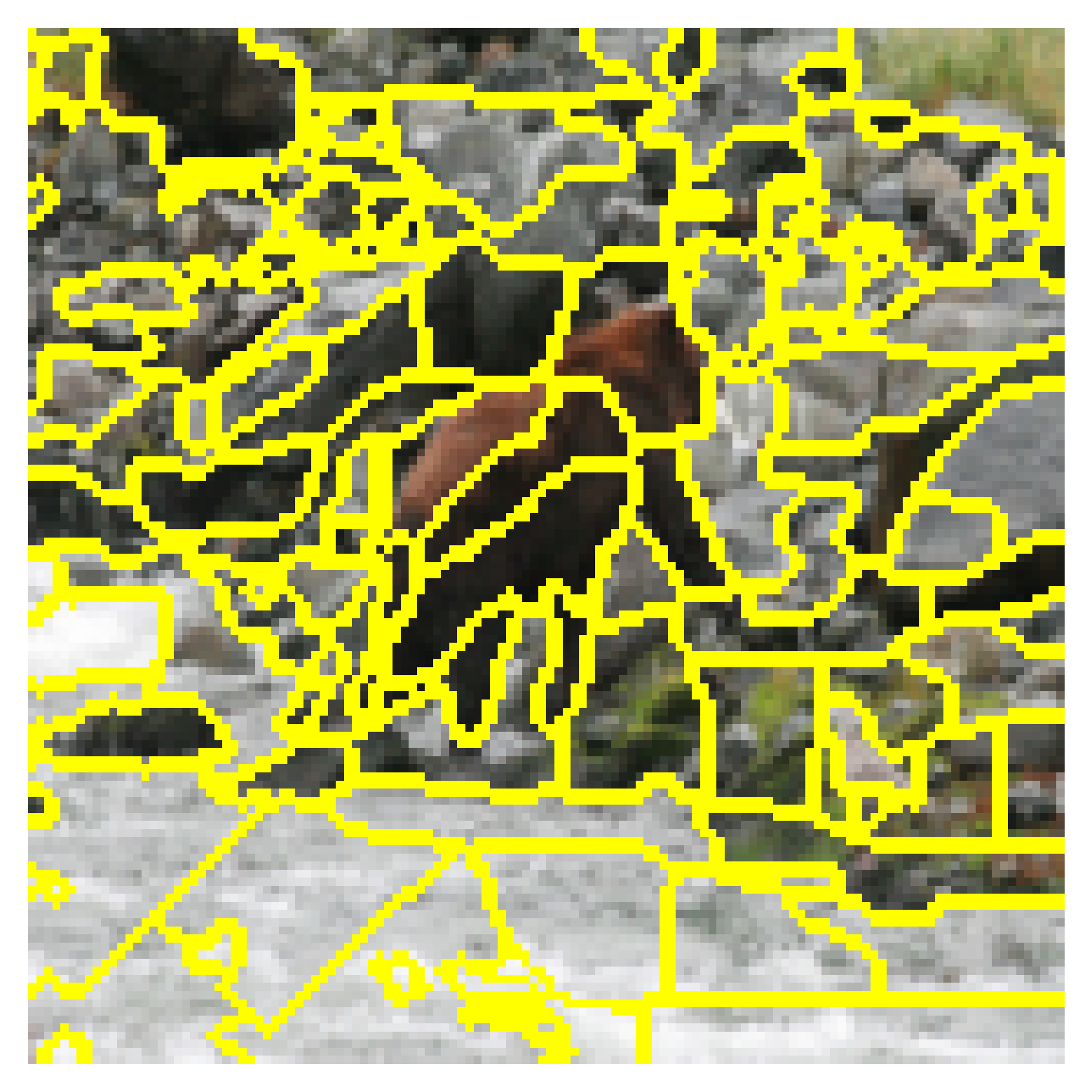}

        \caption{}
    \end{subfigure}
    \begin{subfigure}{0.115\textwidth}
    \includegraphics[width=1\textwidth]{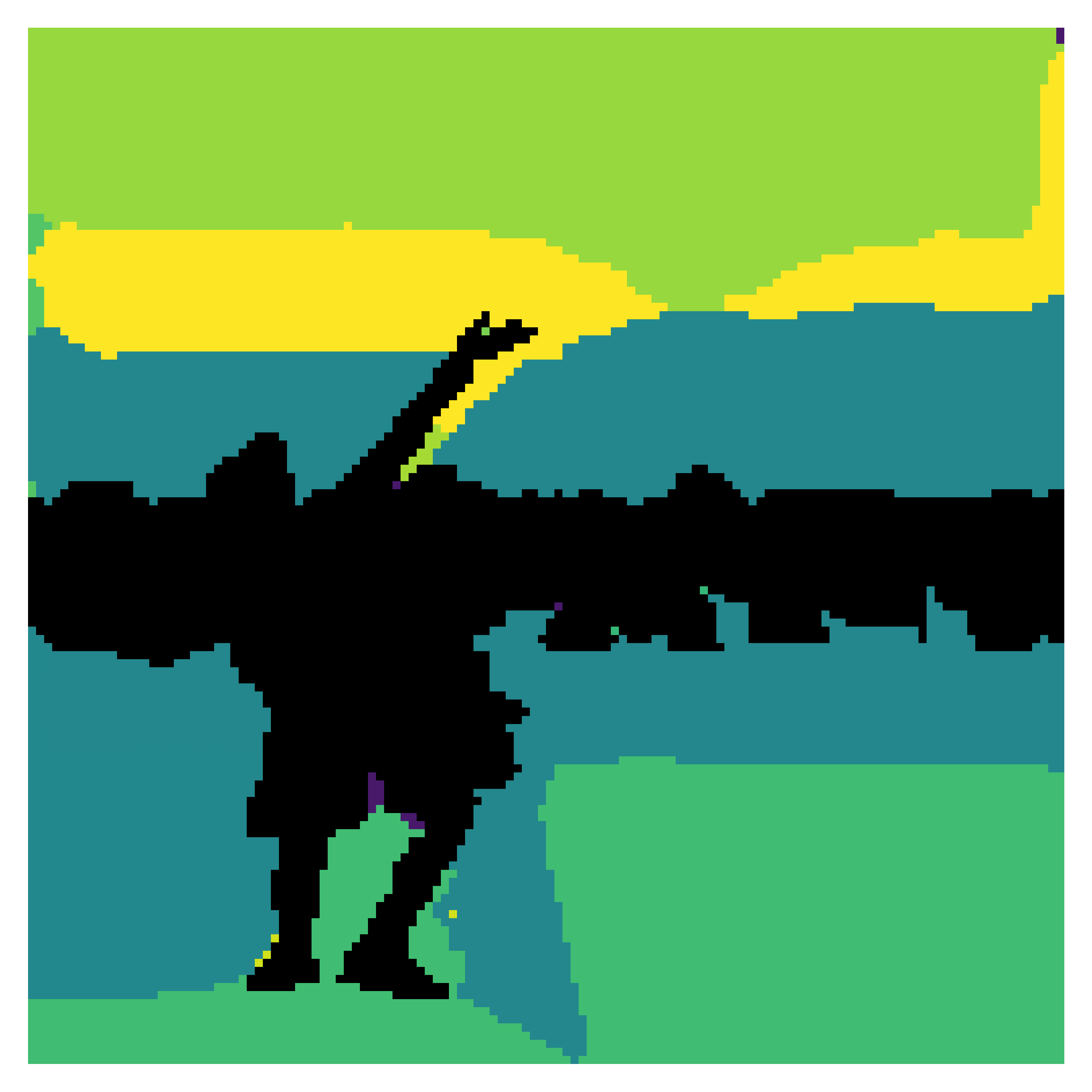} \\
    \includegraphics[width=1\textwidth]{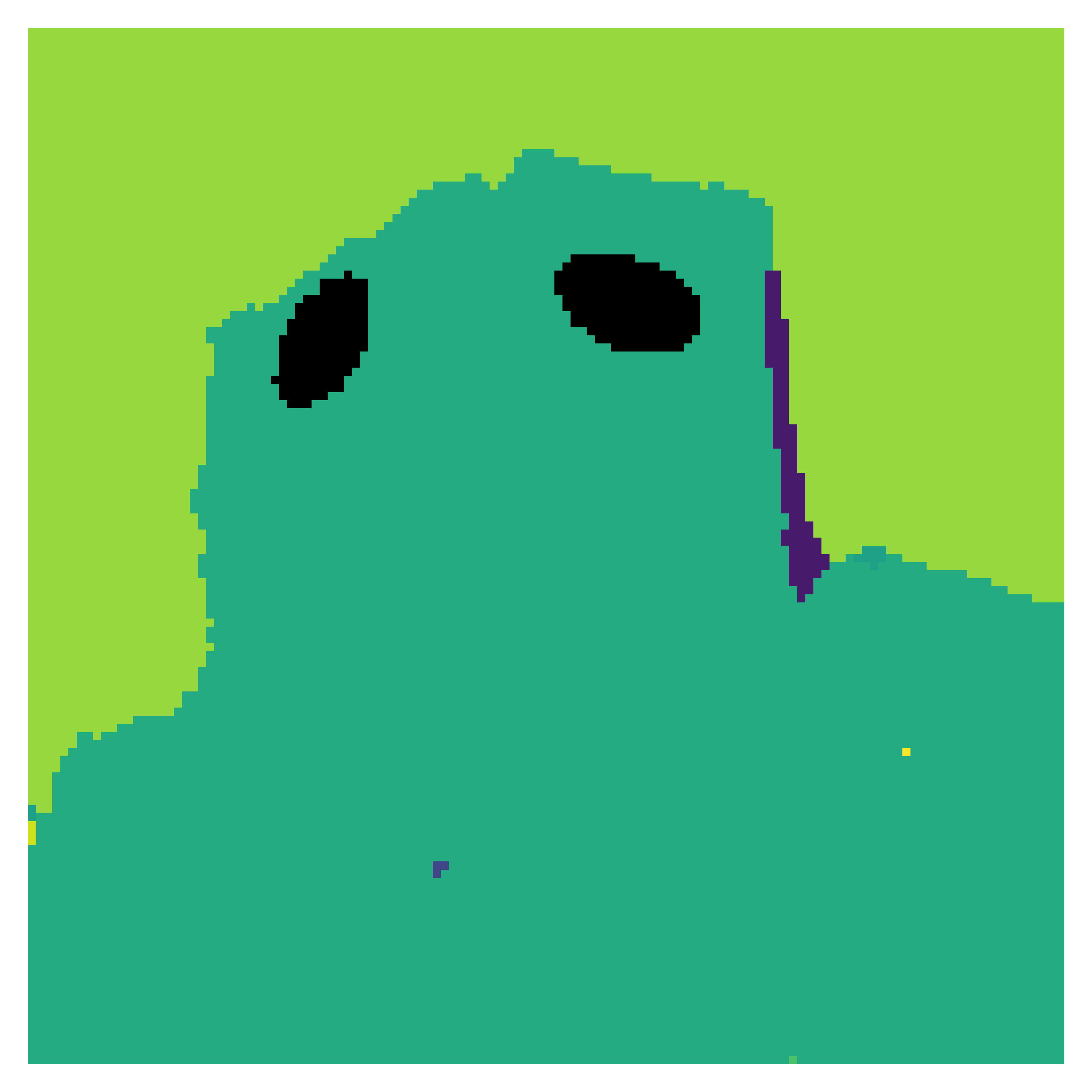}\\
    \includegraphics[width=1\textwidth]{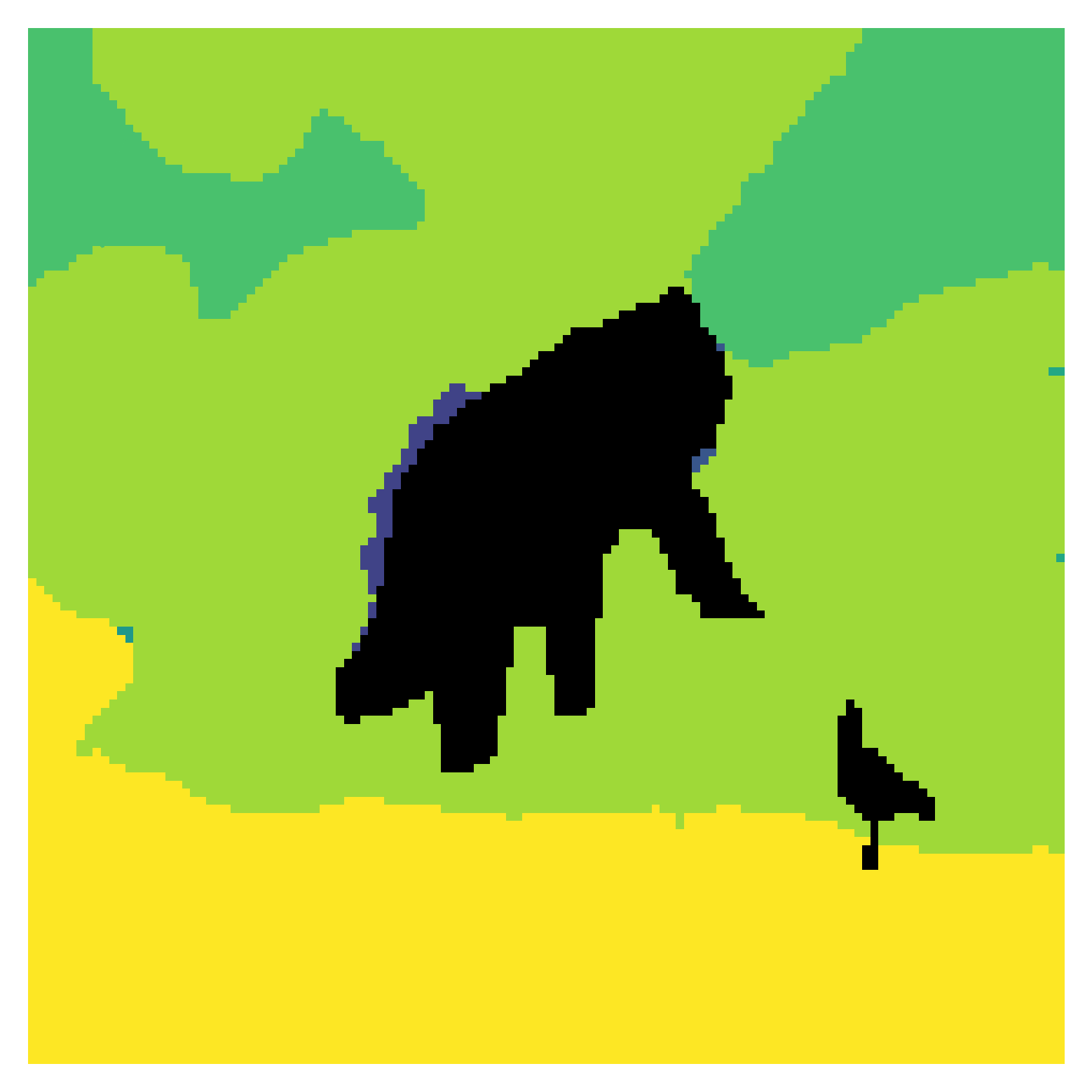}
        \caption{}
    \end{subfigure}
        \begin{subfigure}{0.115\textwidth}
    \includegraphics[width=1\textwidth]{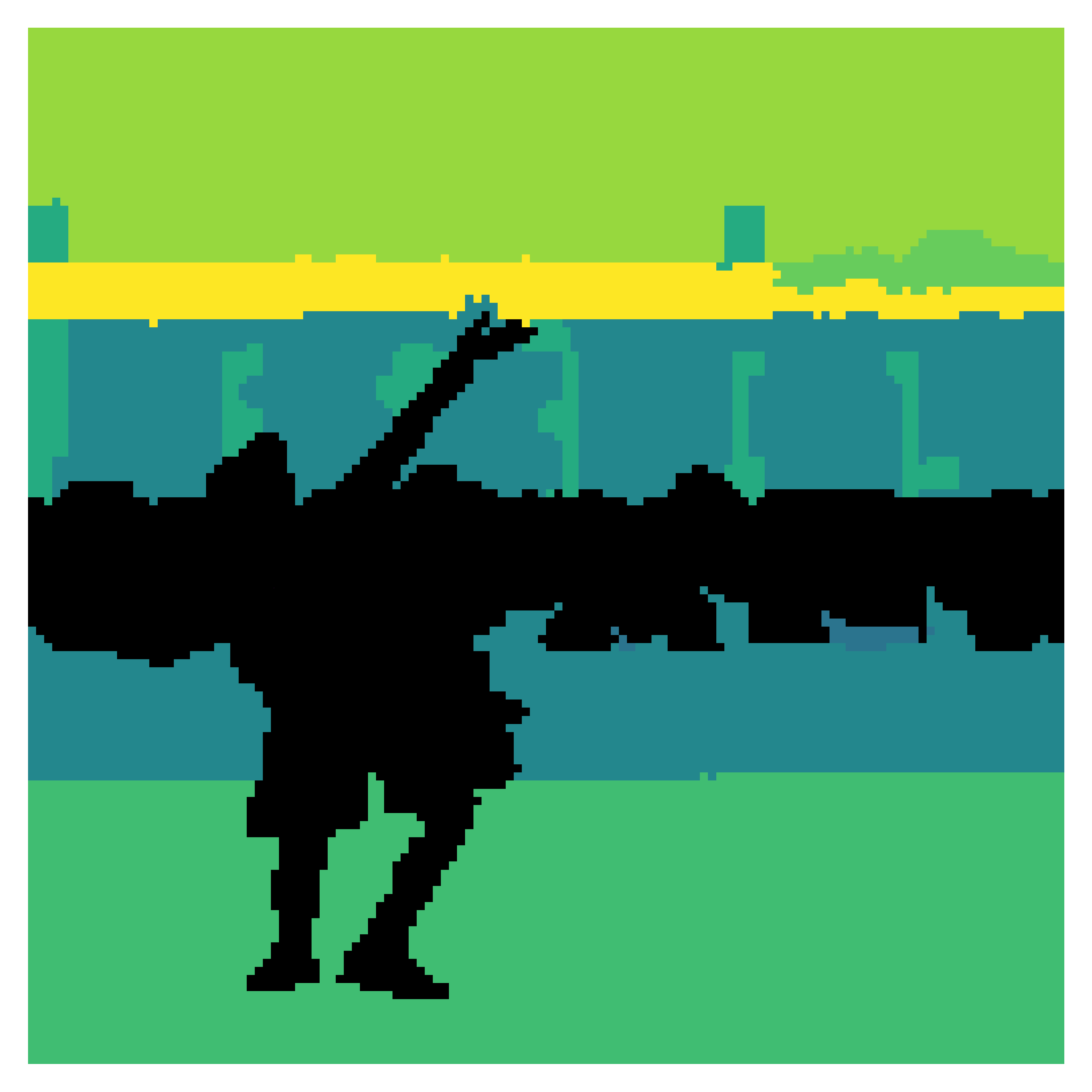} \\
    \includegraphics[width=1\textwidth]{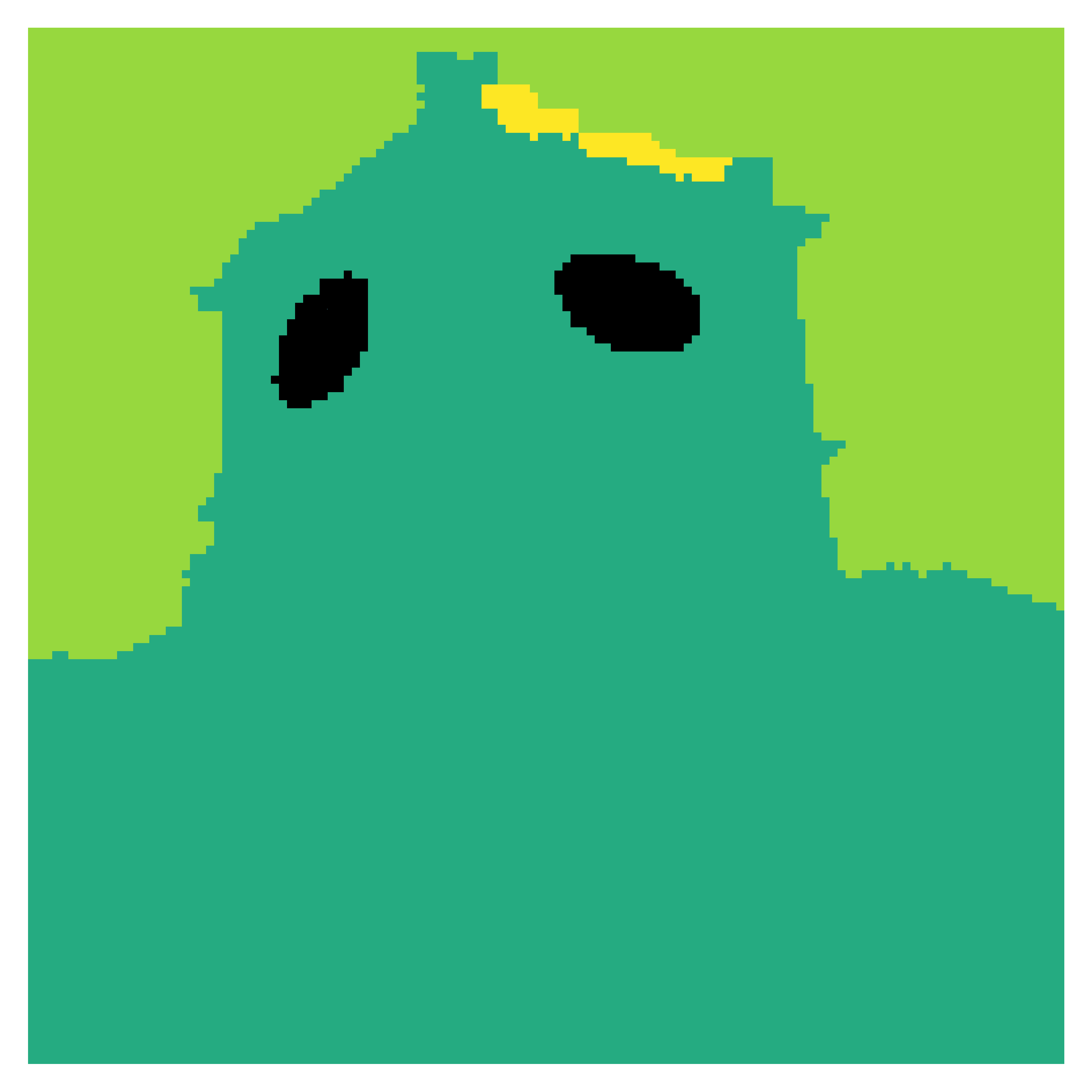}\\
    \includegraphics[width=1\textwidth]{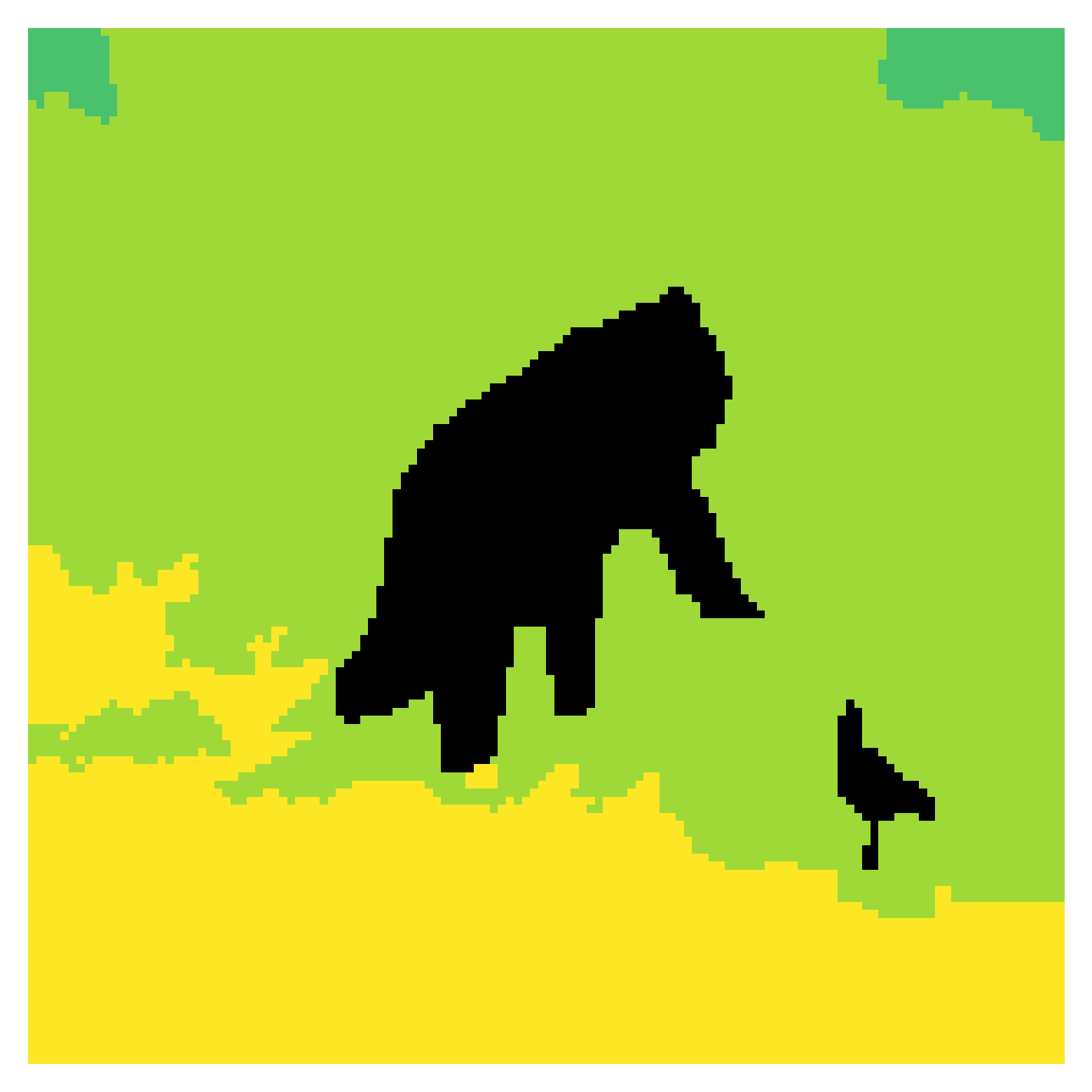}
        \caption{}
    \end{subfigure}
    \caption{Illustration of the superpixel extraction and semantic segmentation on COCO-Stuff. (a) Input images. (b) Superpixelated images. (c) Predicted semantic segmentation maps. (d) Ground-truth  semantic segmentation maps. Non-stuff examples are marked in black.}
    \label{fig:introFig}
\end{figure}

The emergence of Convolutional Neural Networks (CNNs) in recent years \cite{krizhevsky2012imagenet,he2016deep} has tremendously transformed the handling and processing of information in various fields, from Computer Vision \cite{ronneberger2015u, he2016deep, chen2017rethinking} to Computational Biology \cite{senior2020improved} and others. Specifically, the task of supervised semantic segmentation has been widely studied in a series of works like VGG \cite{simonyan2014very}, U-net \cite{ronneberger2015u}, DeepLab \cite{chen2017rethinking} and others \cite{8953615, lin2016refinenet, zhao2017pyramid, fu2019dual, minaee2020image}. However, the task of \emph{unsupervised} image semantic segmentation using deep learning frameworks, where no labels are available has been less researched until recently \cite{ji2019invariant, ouali2020autoregressive, harb2021infoseg, cho2021picie}. Thse methods mostly rely on the concept of Mutual Information Maximization (MIM) which was used for image and volume registration in classical methods \cite{wells1996multi, viola1997alignment}, and recently was incorporated in CNNs and GNNs by \cite{hjelm2018learning} and \cite{deepgraphInfomax}, respectively. The concept of MIM suggests to generate two or more perturbations of the input image (e.g., by geometrical or photometeric variations), feed them to a neural network and demand consistent outputs. The motivation of this approach is to gravitate the network towards learning \emph{semantic} features from the data, while ignoring low-level variations like brightness or affine transformations of the image.

In this paper we propose a novel method that leverages the recent advances in the field of Neural Superpixel Segmentation and GNNs to improve segmentation accuracy. Specifically, given an input image, we propose to first predict its \emph{soft} superpixel representation using a superpixel extraction CNN (SPNN). Then, we employ a GNN to refine and allow the interaction of superpixel features. This step is crucial to share semantic information across distant pixels in the original image. Lastly, we project the learnt features back into an image and feed it to a segmentation designated CNN to obtain the final semantic segmentation map. We distinguish our work from existing methods by first observing that usually other methods rely on a pixel-wise label prediction, mostly varying by the training procedure or architectural changes. Also, while some recent works \cite{mirsadeghi2021unsupervised} propose the incorporation of superpixel information for the task of unsupervised image semantic segmentation, it was obtained from a non-differentiable method like SLIC \cite{slic} or SEEDS \cite{seeds} that rely on low-level features like color and pixel proximity. Our proposed method is differentiable and can be trained in an end-to-end manner, which allows to \emph{jointly} learn the superpixel representation and their high-level features together with the predicted segmentation map. Also, the aforementioned methods did not incorporate a GNN as part of the network architecture. We show in our experiments that those additions are key to obtain accuracy improvement. Our contributions are as follows:
\begin{itemize}
    \item We propose SGSeg: a method that incorporates superpixel segmentation, GNNs and MIM for unsupervised image semantic segmentation.
    \item We show that our network improves the CNN baseline and other recent models, reading similar or better accuracy than state-of-the-art models on 4 datasets.
    \item Our extensive ablation study reveals the importance of superpixels and GNNs in unsupervised image segmentation, and provides a solid evaluation of our method.
\end{itemize}

The rest of the paper is outlined as follows: In Sec. \ref{sec:related} we cover in detail existing methods and related background material. In Sec. \ref{sec:method} we present our method and in Sec. \ref{sec:experiments} we present our numerical experiments.

\section{Related Work}
\label{sec:related}

\subsection{Unsupervised Semantic Segmentation} The unsupervised image segmentation task seeks to semantically classify each pixel in an image without the use of ground-truth labels. Early models like the Geodesic Active Contours \cite{caselles1997geodesic} propose a variational based approach that minimizes a functional to obtain background-object segmentation. Other works propose to extract information from low-level features, for instance by considering the histogram of the red-green-blue (RGB) values of the image pixels \cite{puzicha1999histogram} and by employing a Markov random field \cite{deng2004unsupervised} to model the semantic relations of pixels. In the context of deep learning frameworks, there has been great improvement in recent years \cite{ouali2020autoregressive, harb2021infoseg, ji2019invariant, cho2021picie}. The common factor of those methods is the incorporation of the concept MIM, which measures the similarity between two tensors of possibly different sizes and from different sources \cite{wells1996multi}. Specifically, the aforementioned methods promote the network towards predicting consistent segmentation maps with respect to image transformations and perturbations such as affine transformations \cite{ji2019invariant}. Other methods propose to apply the transformation to the convolutional kernels \cite{ouali2020autoregressive} and demand consistent predictions upon different convolution kernels rasterizations. Such an approach improves the consistency of the learnt features under transformations while avoiding forward and inverse transformation costs in \cite{ji2019invariant}. Additional works like \cite{mirsadeghi2021unsupervised} also follow the idea of MIM but instead of using only geometrical transformations, it is proposed to also employ adversarial perturbations to the image.

Operating on a pixel-level representation of images to obtain a segmentation map is computationally demanding \cite{ke2017multigrid, eliasof2020multigrid}, and modelling the interactions between far pixels in the image requires very deep networks which are also harder to train \cite{haber2019imexnet}. It is therefore natural to consider the compact superpixel representation of the image. The superpixel representation stems from the over-segmentation problem \cite{slic}, where one seeks to segment the image into semantically meaningful regions represented by the superpixels. Also, superpixels are often utilized as an initial guess for non-deep learning based unsupervised image semantic segmentation \cite{xingIcip2016,semanticCRF2018Superpixels,xu2020refining} that greatly reduces the complexity of transferring an input image to its semantic segmentation map. In the context of CNNs, it was proposed \cite{mirsadeghi2021unsupervised} to utilize a superpixel representation based on SLIC in addition to the pixel representation to improve accuracy.

Furthermore, other works propose to use pre-trained networks in order to achieve semantic information of the images, \cite{hamilton2022unsupervised, caron2021emerging}, and \cite{ke2022unsupervised} utilized co-segmentation of from multiviews of images. In this work we focus on networks that are neither trained nor pre-trained with any labelled data and are based on a single image view.

\subsection{Neural Superpixel Segmentation}
The superpixel segmentation task considers the problem of over-segmenting an image, i.e., dividing the image into several sub-regions, where each sub-region has similar features within its pixels. That is, given an image $\bfI \in \mathbb{R}^{H \times W}$, a superpixel segmentation algorithm returns an assignment matrix $\pi \in [0, \ldots, N-1]^{H \times W}$ that classifies each pixel into a superpixel, where $N$ is the number of superpixels. Classical methods like SLIC \cite{slic} use Euclidean coordinates and color space similarity to define superpixels, while other works like SEEDS \cite{seeds} and FH \cite{graph_sp} define an energy functional that is minimized by graph cuts. Recently, methods like \cite{ssn2018} proposed to use CNNs to extract superpixels in a \emph{supervised} manner, and following that it was shown in \cite{suzuki2020superpixel, edgeAwareUnsupervised2021, eliasof2022rethinking} that a CNN is also beneficial for \emph{unsupervised} superpixel extraction and to substantially improve the accuracy of classical methods like SLIC, SEEDS and FH. We note that in addition to performing better than classical methods, the mentioned CNN based models are fully-differentiable, which is a desired property that we leverage in this work. Specifically, it allows the end-to-end learning of superpixels and semantic segmentation from images, which are intimately coupled problems \cite{ssn2018}.

\subsection{Graph Neural Networks} Graph Neural Networks (GNNs) are a generalization of CNNs, operating on an unstructured grid, and specifically on data that can be represented as a graph, like point clouds \cite{wang2018dynamic, eliasof2020diffgcn}, social networks \cite{kipf2016semi} and protein structures \cite{senior2020improved}. Among the popular GNNs one can find ChebNet \cite{defferrard2016convolutional}, GCN \cite{kipf2016semi}, GAT \cite{velickovic2018graph} and others \cite{bruna2013spectral, hamilton2017inductive, xu2018how}. For a comprehensive overview of GNNs, we refer the interested reader to \cite{bronstein2021geometric}.

Building on the success of GNNs in handling sparse point-clouds in 3D, we propose to treat the obtained superpixel representation as a point-cloud in 2D, where each point corresponds to a superpixel located in 2D inside the image boundaries. We may also attach a high-dimensional feature vector to each superpixel, as we elaborate later in Sec. \ref{sec:method}.
We consider three types of GNNs that are known to be useful for data that is geometrically meaningful like superpixels. We start from a baseline PointNet \cite{qi2017pointnet}, which acts as a graph-free GNN consisting of point-wise $1 \times 1$ convolutions. We then examine the utilization of DGCNN \cite{wang2018dynamic} which has shown great improvement over PointNet for 3D points-cloud tasks. However, as we show in Sec. \ref{sec:method}, DGCNN does not consider variable distances of points, which may occur when considering superpixels. We therefore turn to a GNN that considers the distances along the $x$- and- $y$ axes, based on DiffGCN \cite{eliasof2020diffgcn}.

\subsection{Mutual Information in Neural Networks}
The concept of Mutual Information in machine learning tasks has been utilized for image and volume alignment \cite{wells1996multi, viola1997alignment}. Recently, it was implemented into CNNs by the seminal Deep InfoMax \cite{hjelm2018learning} and in GNNs by \cite{deepgraphInfomax}, where unsupervised learning tasks are considered by defining a task that is defined by the data. For example, by demanding signal reconstruction and enforcing MIM between inputs and their reconstruction or predictions.
This concept was found to be useful in a wide array of applications, from image superpixel segmentation \cite{suzuki2020superpixel, edgeAwareUnsupervised2021,eliasof2022rethinking}
to unsupervised image semantic segmentation \cite{ji2019invariant,ouali2020autoregressive, mirsadeghi2021unsupervised} to unsupervised graph related tasks \cite{deepgraphInfomax}. 
In this paper we utilize mutual information maximization in a similar fashion to \cite{ouali2020autoregressive} by demanding similar prediction given different rotations and rasterization of the learnt convolution kernels.

\section{Method}
\label{sec:method}

\begin{figure*}
    \centering
    \includegraphics[width=0.99\textwidth]{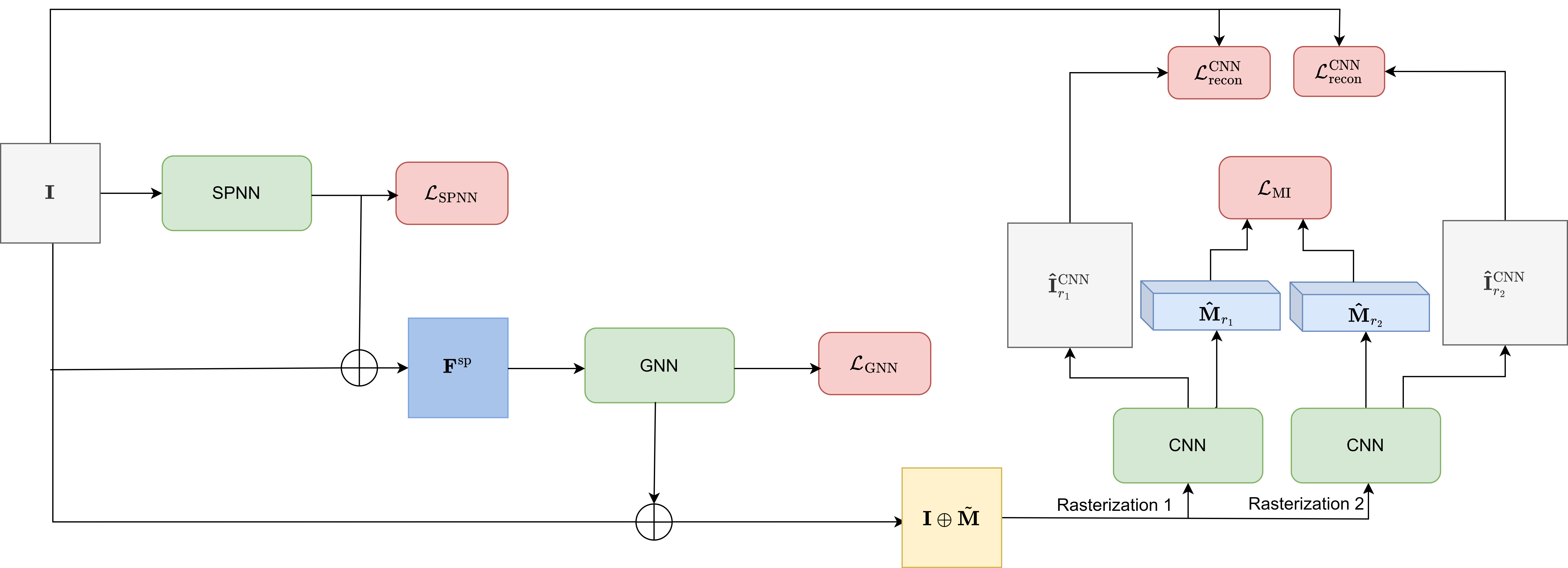}
    \caption{The architecture of SGSeg. $\oplus$ denotes channel-wise concatenation.
    }
    \label{fig:network}
\end{figure*}

We start by defining the notations and setup that will be used throughout this paper.
We denote an RGB input image by $\bfI\in\mathbb{R}^{H\times W\times 3}$, where $H$ and $W$ denote the image height and width, respectively. The goal of the unsupervised image semantic segmentation with $k$ classes is to predict a segmentation map $\hat{\bfM} \in \mathbb{R}^{H \times W \times k}$, and we denote the ground-truth one-hot labels tensor by $\bfM \in \mathbb{R}^{H \times W \times k}$.

\subsection{Superpixel Setup and Notation}
\label{sec:notations}

In this paper we consider superpixels as a medium to predict the desired segmentation map $\hat{\bfM}$.
Let us denote by $N$ the maximal number of superpixels, which is a hyper-parameter of our method. 

In the image superixel segmentation task, the goal is to assign every pixel in the image $\bfI$ to a superpixel. We therefore may treat the pixel assignment prediction as a classification problem. Namely, we denote by
$\bfP\in\mathbb{R}^{H\times W\times N}$ a probabilistic representation of the superpixels. The superpixel to which the $(i, j)$-th pixel belongs is given by the hard-assignment of $s^{i,j} = \mathrm{arg}\max_s\bfP_{i,j,s}$. Thus, we  define value of the $(i, j)$-th pixel of the \textit{hard-superpixelated image} $\bfI^{\bfP}$ as follows:
\begin{equation}
    \label{eq:hardSuperPixImage}
    \bfI^{\bfP}_{i, j} = \frac{\sum_{h,w} \mathbbm{1}_{s^{i,j} = s^{h,w}}\bfI_{h,w}}{\sum_{h,w} \mathbbm{1}_{s^{i,j} = s^{h,w}}},
\end{equation}
where $\mathbbm{1}_{s^{i,j} = s^{h,w}}$ is an indicator function that reads 1 if the hard assignment of the $(i, j)$-th and $(h, w)$-th pixel is the same.
Also, let us define the differentiable \textit{soft-superpixelated image}, which at the $(i, j)$-th pixel reads

\begin{equation}
    \label{eq:softSuperPixImage}
    \hat{\bfI}^{\bfP}_{i, j} = \sum_{s=0}^{N-1} \bfP_{i,j,s} \left(\frac{ \sum_{h,w} \bfP_{h,w,s}\bfI_{h,w}}{\sum_{h,w} \bfP_{h,w,s}} \right).
\end{equation}
Consequently, we can also consider the set of superpixels and their features as a \emph{cloud of points} by denoting $\bfF^{\rm{sp}} = \left[\bfF^{\rm{sp}}_0, \ldots, \bfF^{\rm{sp}}_{N-1}\right]$. This set of points is defined as a weighted average of a features map $\bfF \in \mathbb{R}^{H \times W \times c}$  that includes the $x,y$ coordinates, RGB values and higher dimensional features that stem from the penultimate layer of the SPNN network.
We define the weighting according to $\bfP$ such that feature vector of the $i$-th superpixel is:
\begin{equation}
    \label{eq:superpixelFeature}
    \bfF^{\rm{sp}}_s = \frac{\sum_{h,w} \bfP_{h,w,s}\bfF_{h,w}}{\sum_{h,w} \bfP_{h,w,s}} \in \mathbb{R}^{c}.
\end{equation}

We note that it is also possible to define Eq. \eqref{eq:superpixelFeature} using a hard-assignment as in Eq. \eqref{eq:hardSuperPixImage}. However, such a definition is not differentiable.

\subsection{Semantic Segmentation via Superpixels and GNNs}
\label{sec:methodOverview}
Our network consists of three modules: superpixel extraction neural network (SPNN), a GNN that refines superpixel features, and a final $\rm{CNN}$ segmentation network that predicts class label, per pixel.
Given an image $\bfI$, we first feed it to SPNN to obtain the superpixel assignment map 
\begin{equation}
    \label{eq:superpixelAssignmentExtraction}
    \bfP = \rm{SPNN}(\bfI) \in \mathbb{R}^{H \times W \times N}.
\end{equation} Then, we compute the superpixel features as defined in Eq. \eqref{eq:superpixelFeature} in order to obtain a point cloud $\bfF^{\rm{sp}} \in \mathbb{R}^{N \times c}$ where $c$ is the number of features of each superpixel. We then feed the point cloud $\bfF^{\rm{sp}}$ to the GNN as follows:
\begin{equation}
    \label{eq:pointcloudFeatureRefinement}
    \tilde{\bfF}^{\rm{sp}} = \rm{GNN}(\bfF^{\rm{sp}}) \in \mathbb{R}^{N \times \Tilde{c}},
\end{equation}
where $\tilde{c}$ is the number of output channels of the GNN.

Finally, to obtain an image feature map we project the obtained superpixel features $\tilde{\bfF}^{\rm{sp}}$ to an image using the superpixel assignment map $\bfP$ as follows:
\begin{equation}
    \label{eq:projectionToImage}
     \tilde{\bfM}_{i,j} = \sum_{s=0}^{N-1} \bfP_{i,j,s} \tilde{\bfF}^{\rm{sp}}_{s} \in \mathbb{R}^{\tilde{c}},
\end{equation}
where $\bfP_{i,j,s} \in \mathbb{R}^{H \times W}$ is the probability of $(i,j)$-th pixel to belong the $s$-th superpixel and $\tilde{\bfF}_{s}^{\rm{sp}} \in \mathbb{R}^{\tilde{c}}$ is the feature vector of the $s$-th superpixel.

We then use a segmentation designated CNN fed with a concatenation (denoted by $\oplus$) of the input image $\bfI$ and the projected superpixel features $\tilde{\bfM}$, followed by an application of a SoftMax function to obtain the class probabilities:
\begin{equation}
    \label{eq:finalOutput}
    \hat{\bfM} = \rm{SoftMax}(CNN(\bfI \oplus \tilde{ \bfM})).
\end{equation}

 In what follows we specify the components of the network, namely, $\rm{SPNN}, \rm{GNN}$ and $\rm{CNN}$. The overall architecture is portrayed in Fig. \ref{fig:network}
 
 \subsection{Superpixel Extraction}
 \label{sec:SEmodule}
Our superpixel extraction neural network (SPNN) is based on the proposed methods in \cite{suzuki2020superpixel, edgeAwareUnsupervised2021,eliasof2022rethinking} which were shown to significantly improve the standard metrics of boundary precision and achievable segmentation accuracy compared to methods like SLIC and SEEDS. The core idea is to treat the superpixel segmentation as a classification problem using a CNN. Since no labelled data is available, it is proposed to maximize the mutual information of the superpixels to obtain a deterministic and informative pixel-to-superpixel assignment matrix $\bfP$. In addition, smoothness, reconstruction and edge-awareness losses are utilized to predict a matrix $\bfP$ that is piece-wise constant and adheres to image contours and edges, as desired in the superpixel segmentation problem \cite{slic}. 
Below we present a discussion of the considered losses of SPNN.
\paragraph{SPNN losses.}
We consider the following objectives
\begin{align}
    \nonumber
    \label{eq:allLosses}
    \mathcal{L}_\mathrm{SPNN} =& \mathcal{L}_\mathrm{clustering}+\alpha\mathcal{L}_\mathrm{smoothness}\\ &+ \beta\mathcal{L}_\mathrm{recon} + \eta\mathcal{L}_\mathrm{edge} ,
\end{align}
where $\mathcal{L}_\mathrm{clustering}$ denotes the mutual information loss as presented in ~\cite{suzuki2020superpixel}, based on ~\cite{rim,mut_info}). $\mathcal{L}_\mathrm{smoothness}$ and $\mathcal{L}_\mathrm{recon}$ denote a spatial smoothness and a reconstruction loss, respectively. $\mathcal{L}_\mathrm{edge}$ is an edge-awareness loss that adds prior information to the training process and aims at producing superpixels that follow the edges in the image.
We balance the different terms using $\alpha$, $\beta$ and $\eta$, which are non-negative hyper-parameters.

The objective $\mathcal{L}_\mathrm{clustering}$ is defined as follows:
\begin{align}
\nonumber
    \mathcal{L}_\mathrm{clustering}=&\frac{1}{HW}\sum_{i,j}\sum_{s=0}^{N-1}-\bfP_{i,j,s}\log \bfP_{i,j,s}\\
    &+\lambda\sum_{s=0}^{N-1}\hat{\bfP}_{s}\log\hat{\bfP}_{s},
\end{align}
where $\hat{\bfP}_s=\frac{1}{HW}\sum_{i,j}\bfP_{i,j,s}$ denotes the weight of the $s$-th superpixel.
The first term considers pixel entropy $\bfP_{i,j}\in\mathbb{R}^N$, and promotes the network towards a deterministic superpixel assignment map $\bfP$.
The second term computes the negative entropy of the superpixels weights to obtain weight-uniformity of the superpixels. In our experiments, we set $\lambda = 2$.

Next, the objective $\mathcal{L}_{smoothness}$ measures the difference between neighbouring pixels, by utilizing the popular smoothness prior ~\cite{monodepth, suzuki2020superpixel} that seeks to maximize the correspondence between smooth regions in the image and the predicted assignment map $\bfP$:
\begin{align}
    \nonumber
    \label{eq:basicSmooth}
    \mathcal{L}_\mathrm{smoothness}=\frac{1}{HW}\sum_{i,j}\left(\left\|\partial_x\bfP_{i,j}\right\|_1e^{-\|\partial_x \bfI_{i,j}\|_2^2/\sigma}\right.\\
    +\left.\left\|\partial_y\bfP_{i,j}\right\|_1e^{-\|\partial_y \bfI_{i,j}\|_2^2/\sigma}\right).
\end{align}
Here $\bfP_{i,j}\in\mathbb{R}^N$ and $\bfI_{i,j}$ are the features of the $(i, j)$-th pixel of $\bfP$ and $\bfI$, respectively. $\sigma$ is a scalar set to 10. 

To allow the network to learn image related filters, and since the data is unlabelled, we add 3 additional output channels to SPNN denoted by $\tilde{\bfI} \in \mathbb{R}^{H \times W \times 3}$ to the network, and we minimize the reconstruction objective:
\begin{align}
    \frac{1}{3HW}\sum_{i,j}\|\bfI_{i,j}-\tilde{\bfI}_{i,j}\|_2^2.
\end{align}
We also follow \cite{eliasof2022rethinking} and require the similarity of the predicted soft-superpixelated image  $\hat{\bfI}^{\bfP}$ and the input image $\bfI$. Therefore our total reconstruction loss is given by
\begin{align}
    \label{eq:recons}
    \mathcal{L}_\text{recon}=\frac{1}{3HW} (\sum_{i,j}\|\bfI_{i,j}-\tilde{\bfI}_{i,j}\|_2^2 +  \sum_{i,j}\|\bfI_{i,j}-\hat{\bfI}^{\bfP}_{i,j}\|_2^2).
\end{align}
Lastly, and similar to \cite{edgeAwareUnsupervised2021, eliasof2022rethinking}, we also consider correspondence between the edges of the input image $\bfI$,  the reconstructed image $\tilde{\bfI}$ and the soft-superpixelated image $\hat{\bfI}^{\bfP}$ defined in Eq. \eqref{eq:softSuperPixImage}. This is realized by measuring the Kullback–Leibler (KL) divergence loss, that matches between the edge distributions. The edge maps are computed via the response of the images with a $3\times3$ Laplacian kernel, followed by an application of the SoftMax function to obtain a valid probability function. We denote the edge maps of $\bfI ,\ \tilde{\bfI}, \ \hat{\bfI}^{\bfP}$ by $\rm{E}_\bfI$, $\rm{E}_{\tilde{\bfI}}$ and $\rm{E}_{\hat{\bfI}^{\bfP}}$, respectively.
The edge-awareness loss is defined as follows:
\begin{align}
    \label{eq:edge}
    \mathcal{L}_\text{edge}= \rm{KL}(E_{\bfI}, E_{\tilde{\bfI}}) + \rm{KL}(E_{\bfI}, E_{\hat{\bfI}^{\bfP}}).
\end{align}

\subsection{Graph Neural Network}
\label{sec:GNNmodule}
The superpixels features $\bfF^{\rm{sp}} = \left[\bfF^{\rm{sp}}_0, \ldots, \bfF^{\rm{sp}}_{N-1}\right]$  as defined in Eq. \eqref{eq:superpixelFeature} can be regarded as a small set of points with features. Thus, we suggest to employ a GNN to further process this data and to refine their features. To this end we denote a graph by an ordered tuple $\mathcal{G} = (\mathcal{V},\mathcal{E})$, where $\mathcal{V}=\{0,\ldots, N-1\}$ is a set of superpixel nodes and $\mathcal{E} \subseteq \{(i,j) | i,j \in \mathcal{V} \}$  is a set of edges. To obtain $\mathcal{E}$, we use the $k$-nearest-neighbors ($k$-nn) algorithm with respect to the Euclidean distance of the superpixel features $\bfF^{\rm{sp}}$.

We consider and compare between three GNN backbones for processing the superpixel features. The first is the popular PointNet \cite{qi2017pointnet}, which can be interpreted as a graph-less neural network, where all of the points (superpixels) are lifted into a higher dimension by $1 \times 1$ convolutions and non-linear activations denoted by $h_{\Theta}$ and parameterized by $\Theta$. The feature update rule of PointNet is given by:
\begin{equation}
\label{eq:pointnet}
    \tilde{\bfF}^{\rm{sp}}_{i} = h_{\Theta}(\bfF^{\rm{sp}}_{i}).
\end{equation}
The second considered GNN backbone is DGCNN \cite{wang2018dynamic}, which is widely known for its efficacy in handling geometric point-clouds. It is achieved by
\begin{equation}
\label{eq:edgeconv}
     \tilde{\bfF}^{\rm{sp}}_{i} = \underset{(i,j) \in \mathcal{E}}{\square} h_{\Theta}(g(\bfF^{\rm{sp}}_{i},\bfF^{\rm{sp}}_{j})),
\end{equation}
where $g(\bfF^{\rm{sp}}_{i},\bfF^{\rm{sp}}_{j}) = \bfF^{\rm{sp}}_{i} \oplus (\bfF^{\rm{sp}}_{i} - \bfF^{\rm{sp}}_{j})$, $\square$ is a permutation invariant aggregation operator, and $\oplus$ denotes a channel-wise concatenation operator. For DGCNN, $\square$ is the $\rm{max}$ function.
However, 
as depicted in Fig. \ref{fig:introFig}, the superpixels in the image are not uniformly distributed, i.e.,their center of mass is not uniformly distributed in the image plane. Therefore, DGCNN did not improve beyond the performance of PointNet, as demonstrated in Tab. \ref{tab:GNNcontribution}, 
since DGCNN does not account for the relative distances between points.

\begin{figure}
    \centering
    \begin{subfigure}[t]{0.2\textwidth}
    \includegraphics[width=1\textwidth]{figures/figs1/input.png} \\
    \includegraphics[width=1\textwidth]{figures/figs2/input.png}
    \caption{}
    \end{subfigure}
    \begin{subfigure}[t]{0.2385\textwidth}
    \includegraphics[width=1\textwidth]{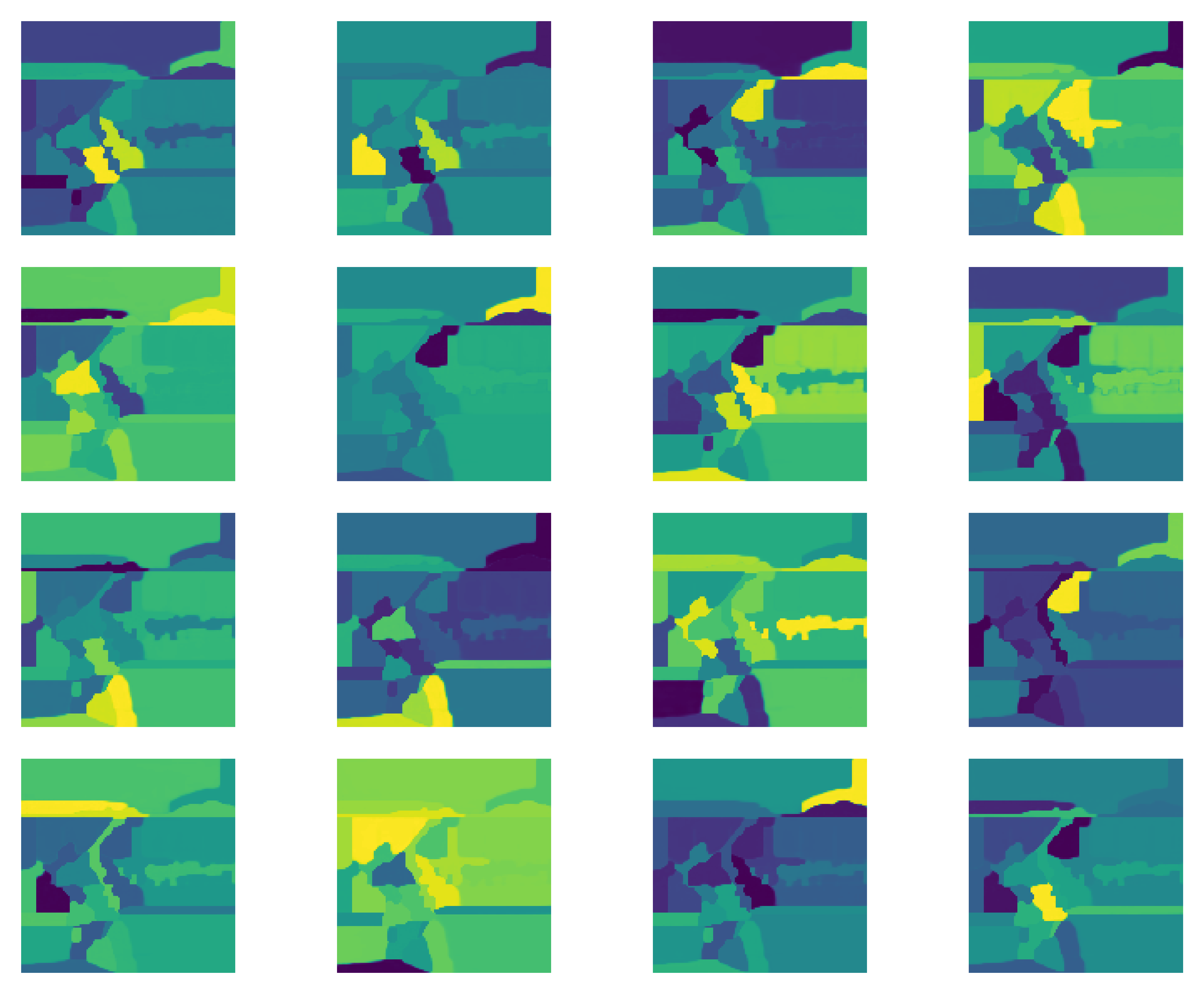} \\
    \includegraphics[width=1\textwidth]{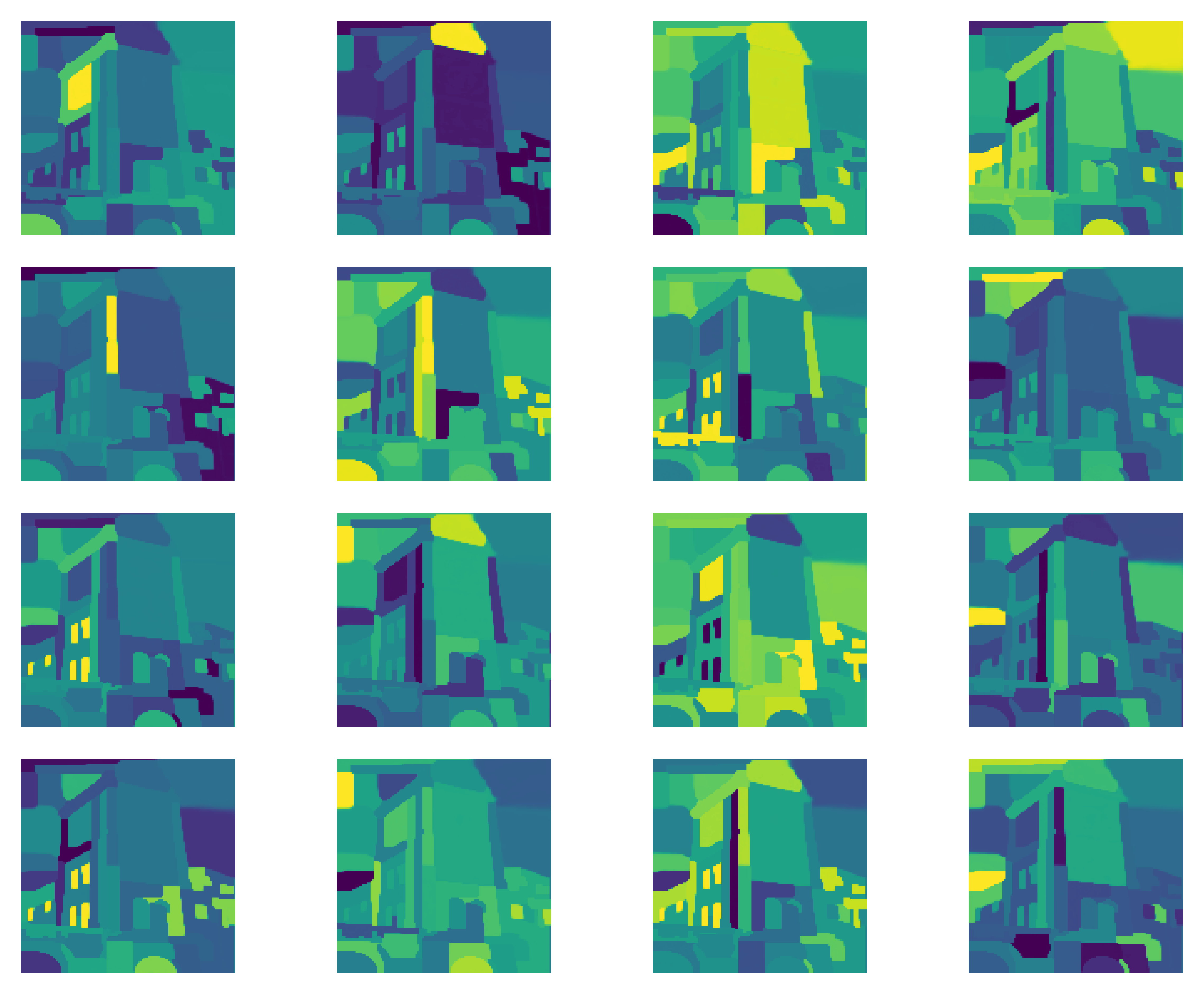}
        \caption{}
    \end{subfigure}
    \caption{(a) Input image. (b) Superpixels features $\tilde{\bfM}$.}
    \label{fig:spFeats}
\end{figure}

To alleviate this problem, we propose a 2D variant of DiffGCN \cite{eliasof2020diffgcn} that offers a directional GNN by replacing $g$ in Eq. \eqref{eq:edgeconv} by 
\begin{equation}
    \label{eq:diffgcn}
    g(\bfF^{\rm{sp}}_{i},\bfF^{\rm{sp}}_{j}) = \bfF^{\rm{sp}}_{i} \oplus (\partial_{x}^{\mathcal{G}}\bfF^{\rm{sp}})_{ij} \oplus (\partial_{y}^{\mathcal{G}}\bfF^{\rm{sp}})_{ij}
\end{equation}
where the partial derivative with respect to the $x$-axis is denoted by $(\partial_{x}^{\mathcal{G}}\bfF^{\rm{sp}})_{ij} = \frac{\bfF^{\rm{sp}}_{i}-\bfF^{\rm{sp}}_{j}}{\Delta(i,j)}(x_i-x_j)$
 and $(\partial_{y}^{\mathcal{G}}\bfF^{\rm{sp}})_{ij}$ is defined analogously for the $y$-axis. $\Delta(i,j)$ denotes the Euclidean distance between the $i$-th and $j$-th superpixels in the image, i.e., $\Delta(i,j) = \sqrt{(x_i-x_j)^2 + (y_i-y_j)^2}.$
We note that for typical 3D point cloud applications like shape classification or segmentation, the formulations of DGCNN and DiffGCN require a coordinate alignment mechanism due to arbitrary rotations that are found in datasets like ShapeNet \cite{chang2015shapenet}. However, since we operate on superpixels that are extracted from 2D images, no further coordinate alignment mechanisms are required.
\paragraph{GNN loss.} The GNN is trained jointly with the semantic segmentation CNN presented in Sec. \ref{sec:CNNmodule}, and is therefore influenced by its losses. In addition, we directly impose a total-variation (TV) loss \cite{rudin1992nonlinear} that promotes smooth and edge-aware features maps of the GNN image-projected output $\tilde \bfM$ (following Eq. \eqref{eq:projectionToImage}). 
More precisely, we minimize the following anisotropic TV loss:
\begin{equation}
    \label{eq:TVloss_GNN}
    \mathcal{L}_\mathrm{TV} = \frac{1}{HW}\sum_{i,j}\left( \|\partial_x{\tilde{\bfM}_{i,j}}\|_1 + \|\partial_y{\tilde{\bfM}_{i,j}}\|_1 \right).
\end{equation}
Thus, the GNN loss is $\mathcal{L}_\mathrm{GNN} = \mathcal{L}_\mathrm{TV}$. In Fig. \ref{fig:spFeats} we show an example of images and part of the obtained projected superpixel features. It is noticeable that semantically related pixels have similar features, and that due to the TV and superpixel over-segmentation process, a piece-wise constant feature map is obtained. We note that this loss may also be defined on the superpixel graph $\mathcal{G}$ and the node (superpixel) features $\tilde{\bfF}^{\rm{sp}}$, but it requires a non trivial adaptation of the loss to graphs. We chose \eqref{eq:TVloss_GNN} for simplicity.

\subsection{Semantic Segmentation Network}
\label{sec:CNNmodule}
We now define the final CNN network that is fed with the channel-wise concatenation of the input RGB image together with its image-projected superpixels features obtained from the GNN, as described in Eq. \eqref{eq:projectionToImage}. We follow the AC approach presented in \cite{ouali2020autoregressive},  where different rasterizations and spatial invariances are considered as a mean to obtain feature maps that correspond to translated and rotated images. Let us denote the predicted segmentation map resulting from two convolution rasterizations $r_1,r_2$ by $\hat{\bfM}_{r_1} = \rm{CNN}_{r_1}(\bfI \oplus \tilde{\bfM}), \ \hat{\bfM}_{r_2} = \rm{CNN}_{r_2}(\bfI \oplus \tilde{\bfM})$ .Given the two predicted segmentation maps $\hat{\bfM}_{r_1}, \hat{\bfM}_{r_2}$, we aim to maximize their mutual information by the following:
\begin{equation}
    \label{eq:MI_segmentation}
    \max \rm{MI}(\hat{\bfM}_{r_1},\hat{\bfM}_{r_2}),
\end{equation} 
where MI denotes mutual information and its corresponding loss is defined as follows:
\begin{equation}
    \label{eq:miDefinition}
    \mathcal{L}_{\rm{MI}}(\hat{\bfM}_{r_1},\hat{\bfM}_{r_2}) = H(\hat{\bfM}_{r_1}) - H(\hat{\bfM}_{r_1} | \hat{\bfM}_{r_2}),
\end{equation}
where $\rm{H}(\cdot)$ denotes the argument entropy. 
That is, we wish to maximize the entropy of $\rm{H}(\hat{\bfM}_{r_1})$ and minimize the conditional entropy $\rm{H}(\hat{\bfM}_{r_1} | \hat{\bfM}_{r_2})$. Therefore, the maximization of Eq. \eqref{eq:miDefinition} aims at obtaining two segmentation maps $\hat{\bfM}_{r_1} , \ \hat{\bfM}_{r_2}$ that are similar despite the different convolution rasterization, while requiring the information of each segmentation map to have a high entropy.

In addition, we propose to add a reconstruction loss to the segmentation CNN by adding 3 channels to the output of $\rm{CNN}$ denoted by $\hat{\bfI}^{\rm{CNN}}$ and minimizing for $r_1, r_2$:
\begin{equation}
    \label{eq:cnnReconstruction}
    \mathcal{L}_\mathrm{recon}^{\rm{CNN}} = \| \hat{\bfI}^{\rm{CNN}}_{r_1} - \bfI  \|_2^2 + \| \hat{\bfI}^{\rm{CNN}}_{r_2} - \bfI  \|_2^2.
\end{equation}

Similarly to the reconstruction loss of $\rm{SPNN}$ in Eq. \eqref{eq:recons}, this objective encourages the network to learn image related filters.
We conclude the overall loss of the semantic segmentation CNN module by:
\begin{equation}
    \label{eq:totalCNNloss}
    \mathcal{L}_\mathrm{CNN} = \mathcal{L}_\mathrm{MI} + \mathcal{L}_\mathrm{recon}^\mathrm{CNN}.
\end{equation}

\subsection{Training SGSeg} 
\label{sec:methodTraining}
We propose two possible training schemes. The first scheme is separable: it starts by training the SPNN module and then trains the GNN and CNN networks separately. The second scheme is an end-to-end training of the whole network together. That is possible since all the components of our method are differentiable. We found the end-to-end training to offer better performance as we can see in the results in Sec. \ref{sec:ablation}. The loss functions of each module (SPNN, GNN, CNN) were discussed previously, and in total we aim to minimize the following objective:
\begin{equation}
    \label{eq:totalLoss}
    \mathcal{L} = \mathcal{L}_\mathrm{SPNN} + \mathcal{L}_\mathrm{GNN} + \mathcal{L}_\mathrm{CNN}.
\end{equation}

\section{Experiments}
\label{sec:experiments}
We now report the experiments conducted to verify the contribution of our method. In Sec. \ref{sec:settings} we provide information on the datasets and training details. In Sec. \ref{sec:unsupExperiments} we report and discuss the results obtained using our approach, and in Sec. \ref{sec:ablation} we conduct an ablation study to delve on the individual components and configurations of our network.
\subsection{Settings and datasets}
\label{sec:settings}
We consider 4 popular unsupervised image segmentation datasets: Potsdam and Potsdam3 datasets \cite{potsdam} which are composed of high-resolution RGBIR aerial images, containing 6 and 3 classes, respectively. Each image is of size of 6000 x 6000 and is split into 200x200 patches for training. Note that while throughout Sec. \ref{sec:method} we refer to the input image as a 3 channels (RGB) tensor, for those datasets the input consists of 4 channels. We also employ the COCO-Stuff and COCO-Stuff3 \cite{8578230} which consist of 164,000 images with pixel-level segmentation maps. The former contains 15 coarse stuff labels and the latter contains 4 labels (ground, sky, plants, and other). In both datasets, we use the same pre-processing as in \cite{ji2019invariant,ouali2020autoregressive} and resize the images to a size of $128 \times 128$. We also follow the standard pixel-accuracy metric, that measures the number of correctly labelled pixels divided by the total count of the pixels in the image. It is also important to note that since our method is unsupervised, there is no direct correspondence between the predicted labels and the ground-truth labelled images of the test set. Therefore, and for a fair comparison with existing methods, we utilize the standard procedure from \cite{ji2019invariant, ouali2020autoregressive, cho2021picie} that uses the Hungarian algorithm \cite{kuhn1955hungarian} to solve a linear assignment problem. This assignment is computed using the segmentation maps of all the images in the test set.

Unless otherwise stated, in all experiments we use DiffGCN as the GNN backbone, and we use the $k$-nn algorithm with $k=20$ to generate the superpixel graph $\mathcal{G}$ using the superpixels features $\bfF^{\rm{sp}}$. In our ablation study, we report the obtained accuracy with various values of $k$. In Appendix A we provide a detailed description of the architectures and the hyper-parameters values used in our experiments.
Our method and experiments are implemented using PyTorch \cite{paszke2017automatic} and PyTorch-Geometric \cite{pyg2019}. Our model is trained and evaluated using an Nvidia RTX-3090 with 24GB memory.

\subsection{Unsupervised Image Segmentation}
\label{sec:unsupExperiments}
We compare our SGSeg both with 'classical' and deep-learning based methods, namely, K-Means \cite{NIPS2004_40173ea4}, Doersch \cite{doersch2015unsupervised}, Isola \cite{isola2015learning}, IIC \cite{ji2019invariant}, AC \cite{ouali2020autoregressive}, InMars \cite{mirsadeghi2021unsupervised} and InfoSeg \cite{harb2021infoseg}. 
Our results are summarized in Tab. \ref{tab:mainResults}. We see that on all of the considered datasets, our method obtains state-of-the-art accuracy. On the COCO-Stuff dataset we obtain pixel accuracy that is on-par with the considered methods, positioned at the top two methods. Our results emphasize the importance of incorporating \emph{global} and \emph{distant} information within the image, as was also suggested in InfoSeg. However, we incorporate this information through superpixels and their feature propagation using a GNN, while InfoSeg learns global features per semantic class.

\begin{table}[t]
\centering
\resizebox{0.45\textwidth}{!}{
\begin{tabular}{lcccc}
 \toprule
 \multirow{2}{*}{Dataset} & COCO- & COCO- & Potsdam & Potsdam3\\
 & Stuff & Stuff3 &  & \\
 \midrule
 Random CNN    & 19.4 & 37.3 & 28.3 & 38.2 \\
 K-Means \cite{NIPS2004_40173ea4}
 & 14.1 & 52.2 & 35.3 & 45.7 \\
 Doersch \cite{doersch2015unsupervised}
 & 23.1 & 47.5 & 37.2 & 49.6 \\
 Isola \cite{isola2015learning}
 & 24.3 & 54.0 & 44.9 & 63.9 \\
 IIC \cite{ji2019invariant}
 & 27.7 & 72.3 & 45.4 & 65.1 \\
 AC \cite{ouali2020autoregressive}
 & 30.8 & 72.9 & 49.3 & 66.5 \\
 InMars \cite{mirsadeghi2021unsupervised}
 & 31.0 & 73.1 & 47.3 & 70.1 \\
 InfoSeg \cite{harb2021infoseg}
 & 38.8 & 73.8 & 57.3 & 71.6 \\
 \midrule
 SGSeg (Ours) & \textbf{39.4} & \textbf{74.6} & \textbf{57.7} & \textbf{71.8}\\
 \bottomrule
\end{tabular}}
\caption{Pixel-accuracy (\%) comparison.}
\label{tab:mainResults}
\end{table}

\subsection{Ablation study}
\label{sec:ablation}
In this section we study the different components and hyper-parameters of our approach. First, we gradually add the proposed components of our SGSeg to separately measure their contribution. Next, we study the quantitative impact of the number of superpixels and the obtained performance from different GNN backbones. Lastly, we study the influence of the superpixel graph density on the accuracy.

\paragraph{Training scheme.}
As discussed in Sec. \ref{sec:methodTraining}, our method can trained in various schemes. First, since SGSeg is fully-differentiable, it can be trained in an end-to-end fashion. Here we ask whether the end-to-end training is beneficial, or if it is more suitable to learn the components weights disjointedly or in a pre-trained fashion. The former proposes to learn the SPNN module first, 'freeze' its weights and then train the GNN and CNN modules. The latter proposes to first train SPNN for a number of epochs, and then train the complete network together. We found that a pre-training approach followed by an end-to-end training to yield the highest accuracy, as depicted in Tab. \ref{tab:trainingScheme}. Thus, in all our experiments we followed this training scheme.

\begin{table}
\centering
\begin{center}
\begin{tabular}{lc}
\toprule
 Training scheme  & Pixel-accuracy (\%)  \\
 \midrule
 Disjoint   & 33.7 \\
 End-to-end   & 36.6  \\
 Pre-train SPNN + end-to-end  & 39.4 \\
 \bottomrule
\end{tabular}
\end{center}
\caption{Influence of the training scheme on COCO-Stuff. Disjoint training scheme first trains SPNN module 
and then trains only the GNN and CNN components.}
\label{tab:trainingScheme}
\end{table}

\paragraph{SGSeg components.}
Our method consists of three components -- $\rm{SPNN}$, $\rm{GNN}$ and $\rm{CNN}$). To obtain a better understanding of their impact on the results, we train and test our method using various combinations of the components. We report the results in Tab. \ref{tab:componentContribution}. Specifically, we first train our segmentation $\rm{CNN}$ network as-is, without the proposed additional $\rm{SPNN}$ and $\rm{GNN}$. As we rely on the AC approach \cite{ouali2020autoregressive}, we see that the obtained pixel-accuracy is similar to the baseline method (our $\rm{CNN}$ obtains 31.1\% vs. 30.8\% of \cite{ouali2020autoregressive}). We note that our $\rm{CNN}$ component is slightly different in architecture and also incorporates an additional reconstruction loss as discussed in Sec. \ref{sec:method}, which also positively contributes to the achieved accuracy. In Appendix A we specify the exact architecture.
Next, we evaluate the performance of our method when adding only the $\rm{SPNN}$ module. This means that we add superpixel information to the $\rm{CNN}$ module, as described in Sec. \ref{sec:methodOverview}, while skipping the GNN step (i.e., skipping Eq. \eqref{eq:pointcloudFeatureRefinement} by setting $\tilde{\bfF}^{\rm{sp}} = \bfF^{\rm{sp}}$). We immediately see an accuracy improvement of 4.6\%, showing the significance of superpixel information.
Finally, by considering our full model that includes both the $\rm{SPNN}$ and $\rm{GNN}$ components, a further accuracy gain of 2.6\% is obtained.

\begin{table}
\centering
\begin{tabular}{ccc}
\toprule
 SPNN & GNN & Pixel-accuracy (\%)\\
 \midrule
  --  & -- &  31.1 \\
  \checkmark  & -- &  35.7 \\
 \checkmark   & \checkmark &  39.4 \\
  \bottomrule
\end{tabular}
\caption{Contribution of the SPNN and GNN components to the pixel-accuracy (\%) on COCO-Stuff.}
\label{tab:componentContribution}
\end{table}

\paragraph{Number of superpixels.} 
The superpixel extraction task demands the adherence of the superpixels to edges and boundaries in the input image and local smoothness, to maximize the achievable-segmentation-accuracy and boundary recall metrics\footnote{See \cite{suzuki2020superpixel} for definitions.}. Thus, as $N$ is decreased, a coarser segmentation of the objects in the image is reflected in the superpixel features, as observed in Fig. \ref{fig:spFeats}.
We therefore study the influence of the number of superpixels on COCO-Stuff and COCO-Stuff3, as reported in Tab. \ref{tab:numSuperpixel}. We find that more superpixels are beneficial when more segmentation classes exist in the dataset. However, increasing $N$ further than 200 harms accuracy, as it reduces the segmentation effect portrayed in Fig. \ref{fig:spFeats}.

\begin{table}
\centering
\begin{tabular}{ccc}
\toprule
  \multirow{2}{*}{\#Superpixels} & \#Segmentation & \multirow{2}{*}{Pixel-accuracy (\%)}\\
  & classes &  \\
 \midrule
  50  & 4 &  73.4 \\
  50  & 15 &  36.7 \\
  \midrule
 100   & 4 &  74.6 \\
  100  & 15 &  37.9 \\
 \midrule
 200   & 4 & 74.5 \\
  200  & 15 & 39.4 \\
   \midrule
 400   & 4 & 74.0 \\
  400  & 15 & 38.5 \\
 \bottomrule
\end{tabular}
\caption{Model hyper-parameters comparison for different number of superpixels and segmentation classes.}
\label{tab:numSuperpixel}
\end{table}

\paragraph{GNN backbone.}
We consider three GNN backbones. The simplest is PointNet \cite{qi2017pointnet} which acts as a graph-free method, and utilizes a series of MLPs to filter the data. We also consider DGCNN \cite{wang2018dynamic} as is well-known and effective for handling 3D point clouds. However, as discussed in Sec. \ref{sec:GNNmodule}, DGCNN was shown to be effective on \emph{uniformly} distributed data points. However, as shown in Fig. \ref{fig:introFig}, superpixels are usually unevenly distributed in the image. We therefore use a 2D variant of DiffGCN \cite{eliasof2020diffgcn}, as explained and motivated in Sec. \ref{sec:GNNmodule}. To study the benefit of the DiffGCN, we compare the achieved accuracy on COCO-Stuff using the three GNN backbones, and report the results in Tab. \ref{tab:GNNcontribution}. We see that while PointNet and DGCNN improve the baseline accuracy from 35.7\% to 36.2\% and 36.0\%, respectively, DiffGCN reads a higher accuracy of 39.4\%.
\begin{table}
\centering
\begin{tabular}{cc}
\toprule
 GNN Backbone & Pixel-accuracy (\%)\\
 \midrule
 Baseline (no GNN) &  35.7 \\
 \midrule
  PointNet  & 36.2 \\
  DGCNN & 36.0 \\
  DiffGCN & 39.4\\
 \bottomrule
\end{tabular}
\caption{GNN backbone influence on the pixel-accuracy on COCO-Stuff.}
\label{tab:GNNcontribution}
\end{table}

\paragraph{Graph density.}
We now turn to study the impact of the graph density. Recall that as a graph we consider the set of superpixels $\bfF^{\rm{sp}}$ as a set of points (i.e., the nodes of the graph), and to obtain the graph edges we use a $k$-nn algorithm which controls the graph edge density. In this experiment we study the effect this hyper-parameter on the obtained accuracy. To this end we examine our SGSeg with $k \in [5,10,20,40]$ and report the obtained pixel-accuracy in Tab. \ref{tab:graphSizeInfluence}. We see that peak accuracy on COCO-Stuff is achieved with $k=20$. We therefore chose this value for the rest of our experiments.
\begin{table}
\centering
\begin{tabular}{cc}
\toprule
 $k$ & Pixel-accuracy (\%)\\
 \midrule
  5  & 36.5 \\
  10 & 37.4 \\
  20 & 39.4\\
40 & 39.0 \\
 \bottomrule
\end{tabular}
\caption{The influence of $k$ using DiffGCN on COCO-Stuff.}
\label{tab:graphSizeInfluence}
\end{table}

\section{Conclusion}
\label{sec:conclusion}
A novel approach that combines Mutual Information Maximization, Neural Superpixel Segmentation and Graph Neural Networks is proposed to tackle the problem of unsupervised image semantic segmentation. The superpixel representation of an image encodes the different parts and entities within an image, and this compact representation can be utilized by the incorporation of GNNs to further refine superpixel features. By fusing these features with the original image and employing a common CNN based approach, a significant improvement is achieved.

\section*{Acknowledgments}
The research reported in this paper was supported by the Israel Innovation Authority through Avatar
consortium, grant no. 2018209 from the United States - Israel
Binational Science Foundation (BSF), Jerusalem, Israel and by the Israeli Council for Higher Education (CHE) via the Data Science Research Center. ME
is supported by Kreitman High-tech scholarship.

{\small
\bibliographystyle{ieee_fullname}
\bibliography{egbib}

\begin{thebibliography}{10}\itemsep=-1pt

\bibitem{slic}
Radhakrishna Achanta, Appu Shaji, Kevin Smith, Aurelien Lucchi, Pascal Fua, and
  Sabine S{\"u}sstrunk.
\newblock S{LIC} superpixels compared to state-of-the-art superpixel methods.
\newblock {\em IEEE transactions on pattern analysis and machine intelligence},
  34(11):2274--2282, 2012.

\bibitem{mut_info}
John~S Bridle, Anthony~JR Heading, and David~JC MacKay.
\newblock Unsupervised classifiers, mutual information and'phantom targets.
\newblock In {\em Advances in neural information processing systems}, pages
  1096--1101, 1992.

\bibitem{bronstein2021geometric}
Michael~M Bronstein, Joan Bruna, Taco Cohen, and Petar Veli{\v{c}}kovi{\'c}.
\newblock Geometric deep learning: Grids, groups, graphs, geodesics, and
  gauges.
\newblock {\em arXiv preprint arXiv:2104.13478}, 2021.

\bibitem{bruna2013spectral}
Joan Bruna, Wojciech Zaremba, Arthur Szlam, and Yann LeCun.
\newblock Spectral networks and locally connected networks on graphs.
\newblock {\em arXiv preprint arXiv:1312.6203}, 2013.

\bibitem{8578230}
Holger Caesar, Jasper Uijlings, and Vittorio Ferrari.
\newblock Coco-stuff: Thing and stuff classes in context.
\newblock In {\em 2018 IEEE/CVF Conference on Computer Vision and Pattern
  Recognition}, pages 1209--1218, 2018.

\bibitem{caron2021emerging}
Mathilde Caron, Hugo Touvron, Ishan Misra, Herv{\'e} J{\'e}gou, Julien Mairal,
  Piotr Bojanowski, and Armand Joulin.
\newblock Emerging properties in self-supervised vision transformers.
\newblock In {\em Proceedings of the IEEE/CVF International Conference on
  Computer Vision}, pages 9650--9660, 2021.

\bibitem{caselles1997geodesic}
Vicent Caselles, Ron Kimmel, and Guillermo Sapiro.
\newblock Geodesic active contours.
\newblock {\em International journal of computer vision}, 22(1):61--79, 1997.

\bibitem{chang2015shapenet}
Angel~X Chang, Thomas Funkhouser, Leonidas Guibas, Pat Hanrahan, Qixing Huang,
  Zimo Li, Silvio Savarese, Manolis Savva, Shuran Song, Hao Su, et~al.
\newblock Shapenet: An information-rich 3d model repository.
\newblock {\em arXiv preprint arXiv:1512.03012}, 2015.

\bibitem{chen2017rethinking}
Liang-Chieh Chen, George Papandreou, Florian Schroff, and Hartwig Adam.
\newblock Rethinking atrous convolution for semantic image segmentation.
\newblock {\em arXiv preprint arXiv:1706.05587}, 2017.

\bibitem{cho2021picie}
Jang~Hyun Cho, Utkarsh Mall, Kavita Bala, and Bharath Hariharan.
\newblock Picie: Unsupervised semantic segmentation using invariance and
  equivariance in clustering.
\newblock In {\em Proceedings of the IEEE/CVF Conference on Computer Vision and
  Pattern Recognition}, pages 16794--16804, 2021.

\bibitem{defferrard2016convolutional}
Micha{\"e}l Defferrard, Xavier Bresson, and Pierre Vandergheynst.
\newblock Convolutional neural networks on graphs with fast localized spectral
  filtering.
\newblock In {\em Advances in neural information processing systems}, pages
  3844--3852, 2016.

\bibitem{deng2004unsupervised}
Huawu Deng and David~A Clausi.
\newblock Unsupervised image segmentation using a simple mrf model with a new
  implementation scheme.
\newblock {\em Pattern recognition}, 37(12):2323--2335, 2004.

\bibitem{doersch2015unsupervised}
Carl Doersch, Abhinav Gupta, and Alexei~A Efros.
\newblock Unsupervised visual representation learning by context prediction.
\newblock In {\em Proceedings of the IEEE international conference on computer
  vision}, pages 1422--1430, 2015.

\bibitem{eliasof2020multigrid}
Moshe Eliasof, Jonathan Ephrath, Lars Ruthotto, and Eran Treister.
\newblock Multigrid-in-channels neural network architectures.
\newblock 2020.

\bibitem{eliasof2020diffgcn}
Moshe Eliasof and Eran Treister.
\newblock Diffgcn: Graph convolutional networks via differential operators and
  algebraic multigrid pooling.
\newblock {\em 34th Conference on Neural Information Processing Systems
  (NeurIPS 2020), Vancouver, Canada.}, 2020.

\bibitem{eliasof2022rethinking}
Moshe Eliasof, Nir~Ben Zikri, and Eran Treister.
\newblock Rethinking unsupervised neural superpixel segmentation.
\newblock {\em arXiv preprint arXiv:2206.10213}, 2022.

\bibitem{graph_sp}
Pedro~F Felzenszwalb and Daniel~P Huttenlocher.
\newblock Efficient graph-based image segmentation.
\newblock {\em International journal of computer vision}, 59(2):167--181, 2004.

\bibitem{pyg2019}
Matthias Fey and Jan~E. Lenssen.
\newblock Fast graph representation learning with {PyTorch Geometric}.
\newblock In {\em ICLR Workshop on Representation Learning on Graphs and
  Manifolds}, 2019.

\bibitem{fu2019dual}
Jun Fu, Jing Liu, Haijie Tian, Yong Li, Yongjun Bao, Zhiwei Fang, and Hanqing
  Lu.
\newblock Dual attention network for scene segmentation, 2019.

\bibitem{glorot2010understanding}
Xavier Glorot and Yoshua Bengio.
\newblock Understanding the difficulty of training deep feedforward neural
  networks.
\newblock In {\em Proceedings of the thirteenth international conference on
  artificial intelligence and statistics}, pages 249--256. JMLR Workshop and
  Conference Proceedings, 2010.

\bibitem{monodepth}
Cl{\'e}ment Godard, Oisin Mac~Aodha, and Gabriel~J Brostow.
\newblock Unsupervised monocular depth estimation with left-right consistency.
\newblock In {\em The IEEE Conference on Computer Vision and Pattern
  Recognition}, pages 270--279, 2017.

\bibitem{haber2019imexnet}
Eldad Haber, Keegan Lensink, Eran Treister, and Lars Ruthotto.
\newblock Imexnet a forward stable deep neural network.
\newblock In {\em International Conference on Machine Learning}, pages
  2525--2534. PMLR, 2019.

\bibitem{hamilton2022unsupervised}
Mark Hamilton, Zhoutong Zhang, Bharath Hariharan, Noah Snavely, and William~T.
  Freeman.
\newblock Unsupervised semantic segmentation by distilling feature
  correspondences.
\newblock In {\em International Conference on Learning Representations}, 2022.

\bibitem{hamilton2017inductive}
William~L. Hamilton, Rex Ying, and Jure Leskovec.
\newblock Inductive representation learning on large graphs.
\newblock In {\em NIPS}, 2017.

\bibitem{harb2021infoseg}
Robert Harb and Patrick Kn{\"o}belreiter.
\newblock Infoseg: Unsupervised semantic image segmentation with mutual
  information maximization.
\newblock In {\em DAGM German Conference on Pattern Recognition}, pages 18--32.
  Springer, 2021.

\bibitem{he2016deep}
Kaiming He, Xiangyu Zhang, Shaoqing Ren, and Jian Sun.
\newblock Deep residual learning for image recognition.
\newblock In {\em Proceedings of the IEEE Conference on Computer Vision and
  Pattern Recognition}, pages 770--778, 2016.

\bibitem{hjelm2018learning}
R~Devon Hjelm, Alex Fedorov, Samuel Lavoie-Marchildon, Karan Grewal, Phil
  Bachman, Adam Trischler, and Yoshua Bengio.
\newblock Learning deep representations by mutual information estimation and
  maximization.
\newblock In {\em International Conference on Learning Representations}, 2019.

\bibitem{isola2015learning}
Phillip Isola, Daniel Zoran, Dilip Krishnan, and Edward~H Adelson.
\newblock Learning visual groups from co-occurrences in space and time.
\newblock {\em arXiv preprint arXiv:1511.06811}, 2015.

\bibitem{potsdam}
ISPRS.
\newblock Isprs 2d semantic labeling contest, 2018.

\bibitem{ssn2018}
Varun Jampani, Deqing Sun, Ming-Yu Liu, Ming-Hsuan Yang, and Jan Kautz.
\newblock Superpixel sampling networks.
\newblock In {\em European Conference on Computer Vision (ECCV)}, pages
  352--368. Springer, 2018.

\bibitem{ji2019invariant}
Xu Ji, Joao~F Henriques, and Andrea Vedaldi.
\newblock Invariant information clustering for unsupervised image
  classification and segmentation.
\newblock In {\em Proceedings of the IEEE/CVF International Conference on
  Computer Vision}, pages 9865--9874, 2019.

\bibitem{ke2022unsupervised}
Tsung-Wei Ke, Jyh-Jing Hwang, Yunhui Guo, Xudong Wang, and Stella~X Yu.
\newblock Unsupervised hierarchical semantic segmentation with multiview
  cosegmentation and clustering transformers.
\newblock In {\em Proceedings of the IEEE/CVF Conference on Computer Vision and
  Pattern Recognition}, pages 2571--2581, 2022.

\bibitem{ke2017multigrid}
Tsung-Wei Ke, Michael Maire, and Stella~X Yu.
\newblock Multigrid neural architectures.
\newblock In {\em Proceedings of the IEEE Conference on Computer Vision and
  Pattern Recognition}, pages 6665--6673, 2017.

\bibitem{kingma2014adam}
Diederik~P Kingma and Jimmy Ba.
\newblock Adam: A method for stochastic optimization.
\newblock {\em arXiv preprint arXiv:1412.6980}, 2014.

\bibitem{kipf2016semi}
Thomas~N Kipf and Max Welling.
\newblock Semi-supervised classification with graph convolutional networks.
\newblock {\em arXiv preprint arXiv:1609.02907}, 2016.

\bibitem{rim}
Andreas Krause, Pietro Perona, and Ryan~G Gomes.
\newblock Discriminative clustering by regularized information maximization.
\newblock In {\em Advances in neural information processing systems}, pages
  775--783, 2010.

\bibitem{krizhevsky2012imagenet}
Alex Krizhevsky, Ilya Sutskever, and Geoffrey~E Hinton.
\newblock Imagenet classification with deep convolutional neural networks.
\newblock {\em Advances in neural information processing systems}, 25, 2012.

\bibitem{kuhn1955hungarian}
Harold~W Kuhn.
\newblock The hungarian method for the assignment problem.
\newblock {\em Naval research logistics quarterly}, 2(1-2):83--97, 1955.

\bibitem{lin2016refinenet}
Guosheng Lin, Anton Milan, Chunhua Shen, and Ian Reid.
\newblock Refinenet: Multi-path refinement networks for high-resolution
  semantic segmentation, 2016.

\bibitem{minaee2020image}
Shervin Minaee, Yuri Boykov, Fatih Porikli, Antonio Plaza, Nasser Kehtarnavaz,
  and Demetri Terzopoulos.
\newblock Image segmentation using deep learning: A survey, 2020.

\bibitem{mirsadeghi2021unsupervised}
S~Ehsan Mirsadeghi, Ali Royat, and Hamid Rezatofighi.
\newblock Unsupervised image segmentation by mutual information maximization
  and adversarial regularization.
\newblock {\em IEEE Robotics and Automation Letters}, 6(4):6931--6938, 2021.

\bibitem{ouali2020autoregressive}
Yassine Ouali, C{\'e}line Hudelot, and Myriam Tami.
\newblock Autoregressive unsupervised image segmentation.
\newblock In {\em European Conference on Computer Vision}, pages 142--158.
  Springer, 2020.

\bibitem{paszke2017automatic}
Adam Paszke, Sam Gross, Soumith Chintala, Gregory Chanan, Edward Yang, Zachary
  DeVito, Zeming Lin, Alban Desmaison, Luca Antiga, and Adam Lerer.
\newblock Automatic differentiation in pytorch.
\newblock In {\em Advances in Neural Information Processing Systems}, 2017.

\bibitem{puzicha1999histogram}
Jan Puzicha, Thomas Hofmann, and Joachim~M Buhmann.
\newblock Histogram clustering for unsupervised image segmentation.
\newblock In {\em Proceedings. 1999 IEEE Computer Society Conference on
  Computer Vision and Pattern Recognition (Cat. No PR00149)}, volume~2, pages
  602--608. IEEE, 1999.

\bibitem{qi2017pointnet}
Charles~R Qi, Hao Su, Kaichun Mo, and Leonidas~J Guibas.
\newblock Pointnet: Deep learning on point sets for 3d classification and
  segmentation.
\newblock In {\em Proceedings of the IEEE Conference on Computer Vision and
  Pattern Recognition}, pages 652--660, 2017.

\bibitem{ronneberger2015u}
Olaf Ronneberger, Philipp Fischer, and Thomas Brox.
\newblock U-net: Convolutional networks for biomedical image segmentation.
\newblock In {\em International Conference on Medical image computing and
  computer-assisted intervention}, pages 234--241. Springer, 2015.

\bibitem{rudin1992nonlinear}
Leonid~I Rudin, Stanley Osher, and Emad Fatemi.
\newblock Nonlinear total variation based noise removal algorithms.
\newblock {\em Physica D: Nonlinear Phenomena}, 60(1):259--268, 1992.

\bibitem{senior2020improved}
Andrew~W Senior, Richard Evans, John Jumper, James Kirkpatrick, Laurent Sifre,
  Tim Green, Chongli Qin, Augustin {\v{Z}}{\'\i}dek, Alexander~WR Nelson, Alex
  Bridgland, et~al.
\newblock Improved protein structure prediction using potentials from deep
  learning.
\newblock {\em Nature}, 577(7792):706--710, 2020.

\bibitem{simonyan2014very}
Karen Simonyan and Andrew Zisserman.
\newblock Very deep convolutional networks for large-scale image recognition.
\newblock {\em arXiv preprint arXiv:1409.1556}, 2014.

\bibitem{8953615}
Ke Sun, Bin Xiao, Dong Liu, and Jingdong Wang.
\newblock Deep high-resolution representation learning for human pose
  estimation.
\newblock In {\em 2019 IEEE/CVF Conference on Computer Vision and Pattern
  Recognition (CVPR)}, pages 5686--5696, 2019.

\bibitem{suzuki2020superpixel}
Teppei Suzuki.
\newblock Superpixel segmentation via convolutional neural networks with
  regularized information maximization, 2020.

\bibitem{seeds}
Michael Van~den Bergh, Xavier Boix, Gemma Roig, Benjamin de Capitani, and Luc
  Van~Gool.
\newblock S{EEDS}: Superpixels extracted via energy-driven sampling.
\newblock In {\em European Conference on Computer Vision (ECCV)}, pages 13--26.
  Springer, 2012.

\bibitem{velickovic2018graph}
Petar Veli{\v{c}}kovi{\'{c}}, Guillem Cucurull, Arantxa Casanova, Adriana
  Romero, Pietro Li{\`{o}}, and Yoshua Bengio.
\newblock {Graph Attention Networks}.
\newblock {\em International Conference on Learning Representations}, 2018.

\bibitem{deepgraphInfomax}
Petar Veličković, William Fedus, William~L. Hamilton, Pietro Liò, Yoshua
  Bengio, and R~Devon Hjelm.
\newblock Deep graph infomax.
\newblock In {\em International Conference on Learning Representations}, 2019.

\bibitem{viola1997alignment}
Paul Viola and William~M Wells~III.
\newblock Alignment by maximization of mutual information.
\newblock {\em International journal of computer vision}, 24(2):137--154, 1997.

\bibitem{wang2018dynamic}
Yue Wang, Yongbin Sun, Ziwei Liu, Sanjay~E Sarma, Michael~M Bronstein, and
  Justin~M Solomon.
\newblock Dynamic graph cnn for learning on point clouds.
\newblock {\em arXiv preprint arXiv:1801.07829}, 2018.

\bibitem{wells1996multi}
William~M Wells~III, Paul Viola, Hideki Atsumi, Shin Nakajima, and Ron Kikinis.
\newblock Multi-modal volume registration by maximization of mutual
  information.
\newblock {\em Medical image analysis}, 1(1):35--51, 1996.

\bibitem{xingIcip2016}
Frank~Z. Xing, Erik Cambria, Win-Bin Huang, and Yang Xu.
\newblock Weakly supervised semantic segmentation with superpixel embedding.
\newblock In {\em 2016 IEEE International Conference on Image Processing
  (ICIP)}, pages 1269--1273, 2016.

\bibitem{xu2018how}
Keyulu Xu, Weihua Hu, Jure Leskovec, and Stefanie Jegelka.
\newblock How powerful are graph neural networks?
\newblock In {\em International Conference on Learning Representations}, 2019.

\bibitem{xu2020refining}
Zhiwei Xu, Thalaiyasingam Ajanthan, and Richard Hartley.
\newblock Refining semantic segmentation with superpixel by transparent
  initialization and sparse encoder.
\newblock {\em arXiv preprint arXiv:2010.04363}, 2020.

\bibitem{edgeAwareUnsupervised2021}
Yue Yu, Yang Yang, and Kezhao Liu.
\newblock Edge-aware superpixel segmentation with unsupervised convolutional
  neural networks.
\newblock In {\em 2021 IEEE International Conference on Image Processing
  (ICIP)}, pages 1504--1508, 2021.

\bibitem{NIPS2004_40173ea4}
Lihi Zelnik-manor and Pietro Perona.
\newblock Self-tuning spectral clustering.
\newblock In L. Saul, Y. Weiss, and L. Bottou, editors, {\em Advances in Neural
  Information Processing Systems}, volume~17. MIT Press, 2004.

\bibitem{zhao2017pyramid}
Hengshuang Zhao, Jianping Shi, Xiaojuan Qi, Xiaogang Wang, and Jiaya Jia.
\newblock Pyramid scene parsing network, 2017.

\bibitem{semanticCRF2018Superpixels}
Wei Zhao, Yi Fu, Xiaosong Wei, and Hai Wang.
\newblock An improved image semantic segmentation method based on superpixels
  and conditional random fields.
\newblock {\em Applied Sciences}, 8(5), 2018.

\end{thebibliography}
}

\clearpage

\appendix

\section{Architectures and hyper-parameters}
\label{appendix:architectures}
\subsection{Architectures}
We now elaborate on the specific architectures used in our experiments in Sec. \ref{sec:experiments}. Recall that our method consists of three modules -- $\rm{SPNN}$, $\rm{GNN}$ and $\rm{CNN}$. The overall architecture is given in Fig. \ref{fig:network} in the main paper. In Tab. \ref{table:SPNN}--\ref{table:CNN} we specify the respective components.  All layers are initialized using Xavier \cite{glorot2010understanding}
initialization.
We denote the initial number of features by $c_{in}$.
\paragraph{SPNN Architecture.}
We present the architecture in Tab. \ref{table:SPNN}.
In the case of COCO-Stuff and COCO-Stuff3, the input is an RGB image, and therefore $c_{in}^{\rm{SPNN}}=3$. For Potsdam and Potsdam3 the input is an RGBIR image, thus $c_{in}^{\rm{SPNN}}=4$. We denote a 2D convolution by Conv2D, a 1D convolution by Conv1D. A batch-normalization is denoted by BN. To obtain hierarchical information we follow \cite{eliasof2022rethinking} and employ $3 \times3$ depth-wise Atrous convolution \cite{chen2017rethinking} in our SPNN, with dilation rates of [1,2,4] denoted by DWA [1,2,4]. We then concatenate their feature maps, denoted by $\oplus$. Recall that we demand 3 additional output channels for SPNN to obtain an image reconstruction. We therefore denote the total output number of channels by $\tilde{N} = N + 3$.

\begin{table}[t]
  \small
  \begin{center}
  \resizebox{0.5\textwidth}{!}{
  \begin{tabular}{lcc}
  \toprule
    Input size & Layer  &  Output size \\
    \midrule
    $H \times W \times c_{in}^{\rm{SPNN}}$ & $5\times5$ Conv2D, BN, ReLU & $H \times W \times 64$\\
     $H \times W \times 64$ & $3\times3$ Conv2D, BN, ReLU & $H \times W \times 128$ \\
      $H \times W \times 128$ & $3\times3$ Conv2D, BN, ReLU & $H \times W \times 256$ \\
    $H \times W \times 256$ & $3\times3$ Conv2D, BN, ReLU & $H \times W \times 512$ \\
    $H \times W \times 512$ & $3\times3$ DWA [1,2,4], $\oplus$ & $H \times W \times 1536$ \\
    $H \times W \times 1536$ & $3\times3$ Conv2D, BN, ReLU & $H \times W \times 512$ \\
    $H \times W \times 512$ & $1\times1$ Conv2D & $H \times W \times \tilde{N}$ \\
    \bottomrule
  \end{tabular}}
\end{center}
\caption{SPNN architecture.}
  \label{table:SPNN}

\end{table}

\paragraph{GNN architecture.}
We present the GNN architecture in Tab. \ref{table:GNN}. Since we consider the GNN backbones of PointNet, DGCNN and DiffGCN, we denote a GNN by $\rm{GNN-block}$, and depending on the chosen backbone, it is replaced with its respective network, which is the same as specified in Sec. \ref{sec:method}. The input to the GNN is a concatenation of the $x$ and $y$ coordinates, the mean RGB (RGBIR for Potsdam and Potsdam3)
values that belongs of each superpixel, together with the average superpixel high-level feature that is taken from the penultimate layer of SPNN, which is a vector in $\mathbb{R}^{512}$, as defined in Eq. \eqref{eq:superpixelFeature} in the main paper. Thus, the input to the GNN is of shape $N \times 517$ for COCO-Stuff and COCO-Stuff3, and $N \times 518$ for Potsdam and Potsdam3. We therefore denote the GNN input number of channels by $c_{in}^{\rm{GNN}}$. We perform a total of $L$ GNN-blocks, and we concatenate their feature maps, denoted by $L \times \oplus$.  In our experiments, we set $L=4$. Following that, standard $1\times1$ convolutions are performed to output a tensor with a final shape of $N\times64$. This architecture is based on the general architecture in DGCNN.

\begin{table}[t]
\small
  \begin{center}
  \begin{tabular}{lcc}
  \toprule
    Input size & Layer  &  Output size \\
    \midrule
    $N \times c_{in}^{\rm{GNN}}$ & $1\times1$ Conv1D, BN, ReLU & $N \times 64$ \\
    $N \times 64$ & $L \times $ $\rm{GNN-block}$ & $N \times 64$ \\
    $N \times L\cdot 64$ & $L \times \oplus$ & $N \times L\cdot 64$ \\
    $N \times L\cdot64$ & $1\times1$ Conv1D, BN, ReLU  & $N \times 256$\\
    $N \times 256$ & $1\times1$ Conv1D, BN, ReLU  & $N \times 128$\\
    $N \times 128$ & $1\times1$ Conv1D & $N \times 64$\\
    \bottomrule
  \end{tabular}
\end{center}
\caption{GNN architecture.}
\label{table:GNN}
\end{table}

\paragraph{CNN Architecture.} Our segmentation CNN is based on general architecture of AC. It consists of 4 ResNet\cite{he2016deep} (residual) blocks, where at each block the first convolution is the rasterized AC convolution \cite{ouali2020autoregressive} We specify the
residual block in Tab. \ref{table:residualBlock}. We denote the input number of channels for the residual block by $c$.

\begin{table}[t]
\small
  \begin{center}
  \begin{tabular}{lcc}
  \toprule
    Input size & Layer  &  Output size \\
    \midrule
    $H \times W \times c$ & $3\times3$ AC, BN, ReLU  & $H \times W \times 2c$ \\
    $H \times W \times 2c$ & $1\times1$ Conv2D,BN,ReLU  & $H \times W \times 2c$
    \\
    $H \times W  \times c$ & Input zero-padding  & $H \times W \times 2c$\\
    $H \times W  \times 2c$ & Residual-connection  & $H \times W \times 2c$\\
    \midrule
    $H \times W  \times 2c$ & $1\times1$ Conv2D, BN, ReLU  & $H \times W \times 2c$\\
    $H \times W  \times 2c$ & $1\times1$ Conv2D  & $H \times W \times 2c$\\
        $H \times W  \times 2c$ & $1\times1$ Residual-connection  & $H \times W \times 2c$\\
    \midrule
        $H \times W  \times 2c$ & $1\times1$ Conv2D, BN, ReLU  & $H \times W \times 2c$\\
    $H \times W  \times 2c$ & $1\times1$ Conv2D  & $H \times W \times 2c$\\
        $H \times W  \times 2c$ & $1\times1$ Residual-connection  & $H \times W \times 2c$\\
    \bottomrule
  \end{tabular}
\end{center}
  \caption{The architecture of a residual block. AC denotes the autoregressive, rasterized 2D convolution from \cite{ouali2020autoregressive}.}
    \label{table:residualBlock}
\end{table}

The complete segmentation CNN architecture is given in Tab. \ref{table:CNN}. We feed it with a concatenation of the input image and the projected superpixels features, as described in Eq. \eqref{eq:finalOutput}. As discussed earlier, because COCO-Stuff and COCO-Stuff3 consist of RGB images, while Potsdam and Potsdam3 consist of RGBIR images, we denote the input channels to the CNN by $c_{in}^{\rm{CNN}} = 64 + c_{in}^{\rm{SPNN}}$, where $c_{in}^{\rm{SPNN}} = 3$ for the former and $c_{in}^{\rm{SPNN}}=4$ for the latter datasets. The 64 channels stem from the projected superpixels features $\tilde{\bfM}$ as described in Eq. \eqref{eq:projectionToImage} in the main paper. As discussed in Sec. \ref{sec:CNNmodule}, we demand the segmentation CNN network to also reconstruct the input image. We therefore have $c_{out}^{\rm{CNN}} = k + c_{in}^{SPNN}$ output channels where $k$ is the number of semantic classes. To obtain valid probability distribution of the segmentation prediction, we apply the SoftMax function to the first $k$ features of the CNN output.
To reconstruct the image we apply a further $3 \times 3$ 2D convolution after the bilinear-interpolation step in Tab. \ref{table:CNN}. Note that these two final CNN steps, of SoftMax and Conv2D are not sequential. In particular their input is the output of the bilinear-interpolation step, and their results are the output of the segmentation CNN network. 
\begin{table}[t]
\def\arraystretch{1.03}
\small
  \begin{center}
  \resizebox{0.5\textwidth}{!}{
  \begin{tabular}{lcc}
  \toprule
    Input size & Layer  &  Output size \\
    \midrule
    $H \times W \times  \times c_{in}^{\rm{CNN}}$ & $3 \times 3$ Conv2D, BN, ReLU & $H \times W \times 64$ \\
    $H \times W \times 64$ & $2\times2$ Max-Pooling & $\frac{H}{2} \times \frac{W}{2} \times 64$\\
    $\frac{H}{2} \times \frac{W}{2} \times 64$ & Residual-block & $\frac{H}{2} \times \frac{W}{2} \times 128$ \\
    $\frac{H}{2} \times \frac{W}{2} \times 64$ & Residual-block & $\frac{H}{2} \times \frac{W}{2} \times 128$ \\
    $\frac{H}{2} \times \frac{W}{2} \times 128$ & Residual-block & $\frac{H}{2} \times \frac{W}{2} \times 256$ \\
    $\frac{H}{2} \times \frac{W}{2} \times 256$ & Residual-block & $\frac{H}{2} \times \frac{W}{2} \times 512$ \\
    $\frac{H}{2} \times \frac{W}{2} \times 512$ & $1 \times 1$ Conv2D & $\frac{H}{2} \times \frac{W}{2} \times c_{out}^{\rm{CNN}}$ \\
    $\frac{H}{2} \times \frac{W}{2} \times c_{out}^{\rm{CNN}}$ & $1\times 1$ Conv2D & $\frac{H}{2} \times \frac{W}{2} \times c_{out}^{\rm{CNN}}$ \\
    $\frac{H}{2} \times \frac{W}{2} \times c_{out}^{\rm{CNN}}$ &
    Bilinear-interpolation & $H \times W \times c_{out}^{\rm{CNN}}$ \\
    \midrule
    \multirow{2}{*}{$H \times W \times c_{out}^{\rm{CNN}}$} & SoftMax-$k$ & $H \times W \times k$ \\
    &  $3 \times 3$ Conv2D  & $H \times W \times c_{in}^{\rm{SPNN}}$ \\
    \bottomrule
  \end{tabular}}
\end{center}
  \caption{Segmentation CNN architecture. SoftMax-$k$ applies a SoftMax application to the first $k$ channels. The outputs of the network are the segmentation map and the image reconstruction.}
    \label{table:CNN}
\end{table}

\paragraph{Run-times} We measure the training and inference run-times and accuracy of our method and compare it with the baseline AC \cite{ouali2020autoregressive} that we take inspiration from in our final CNN segmentation block. The run-times are given in Tab. \ref{tab:runtimes}. Clearly, our method requires slightly more computational time, due to the added components of $\rm{SPNN}$ and $\rm{GNN}$ blocks. However, this slight increase in computational cost also returns significantly better pixel-accuracy across all the considered datasets (as can also be seen in Tab. \ref{tab:mainResults}.

\begin{table}[]
    \centering
    \small
    \begin{tabular}{cccc}
    \toprule
        Method &  Train & Inference & Accuracy \\
        \midrule
        AC \cite{ouali2020autoregressive} & 62.81 & 9.89 & 30.8 \\
        SGSeg (Ours) & 76.34 & 12.91 & 39.4 \\
        \bottomrule
    \end{tabular}
    \caption{Run-times [ms] and accuracy (\%) on COCO-Stuff.}
    \label{tab:runtimes}
\end{table}

\subsection{Hyper-parameters}
We now provide the selected hyper-parameters in our experiments. In all experiments, we initialize the weights by the Xavier initialization. We employ the Adam \cite{kingma2014adam} with the default parameters ($\beta_1 = 0.9$ and $\beta_2 = 0.999$). In Tab. \ref{tab:hyperparams} we report the learning-rate and batch size (denoted by BS) and loss balancing terms $\alpha, \beta, \gamma$. The number of superpixels is 200 for COCO-Stuff and 100 for the rest, following the results of our ablation study in Tab. \ref{tab:numSuperpixel} in the main paper. The number of neighbors in our GNN is $k=20$. We pre-train the SPNN component for 10 epochs prior to the end-to-end training, which is identical to \cite{ouali2020autoregressive}, and also uses the same data-augmentation techniques. In our ablation studies, we use the same hyper-parameters unless otherwise reported in the main text, as some of the ablation  studies examine the impact of the hyper-parameters.

We denote the learning rate of our SPNN, GNN and CNN networks by $LR_{\rm{SPNN}}, \ LR_{\rm{GNN}}, \ LR_{\rm{CNN}}$, respectively.
We did not find weight-decay to improve the results, and therefore we set it to 0 in all experiments.
Our hyper-parameters search space in all the experiments is as follows: $LR_{\rm{SPNN}}, \ LR_{\rm{GNN}}, \ LR_{\rm{CNN}}\in [1e-4, 1e-6]$, and $\alpha, \beta, \gamma \in [0.5,10]$ and $\rm{BS} \in \{16,32,64\}$.

\begin{table}[t]
  \begin{center}
  \resizebox{1.0\linewidth}{!}{\begin{tabular}{llcccccccccc}
  \toprule
     Dataset & $LR_{SPNN}$ & $LR_{GNN}$ & $LR_{CNN}$ & \rm{BS} & $\alpha$ & $\beta$ & $\gamma$ \\
    \midrule
    COCO-Stuff & 1e-5 & 5e-4 &  5e-6 &  64 &  2 & 5 & 1 \\
    COCO-Stuff3 & 1e-5 & 5e-4  &  5e-5 & 64  & 2 & 5 & 1   \\
    Potsdam &  5e-5 & 1e-4 & 1e-6 &  32 &  1 & 5 & 0.5  \\
    Potsdam3 & 5e-5 & 5e-4 & 5e-6 &  32 &  1 & 5 & 0.5 \\
    \bottomrule
  \end{tabular}}
\end{center}
  \caption{Hyper-parameters values used in our experiments.}
  \small
  \label{tab:hyperparams}
\end{table}

\end{document}